\documentclass[twoside]{article}

\usepackage{PRIMEarxiv}
\usepackage[utf8]{inputenc} 
\usepackage[T1]{fontenc}    
\usepackage{hyperref}       
\usepackage{url}            
\usepackage{booktabs}       
\usepackage{amsfonts}       
\usepackage{nicefrac}       
\usepackage{microtype}      
\usepackage{lipsum}
\usepackage{fancyhdr}       
\usepackage{graphicx}       
\graphicspath{{media/}}     

\usepackage{lipsum}
\usepackage{float} 
\usepackage{bm}
\usepackage{tikz}
\usetikzlibrary{shapes.geometric, arrows.meta, positioning}

\usepackage{tcolorbox}
\usepackage{listings}
\usepackage{xcolor}
\tcbuselibrary{listings, skins, breakable}

\title{From Handwriting to Structured Data: Benchmarking AI Digitisation of Handwritten Forms 
}

\author{
  Nicholas Pather$^\star$ \\
  CSAM, University of the Witwatersrand \\
  \& Grai Labs \\
  Johannesburg, South Africa \\
  \texttt{pathernicholas@gmail.com} \\
  \And
  Joshua Fouché$^\star$ \\
  Grai Labs \\
  Cape Town, South Africa \\
  \texttt{joshuajdfouche28@gmail.com} \\
  \And
  Sitwala Mundia \\
  Faculty of Health Sciences,\\
  University of the Witwatersrand \\
  South Africa \\
  \texttt{sitwala.mundia@gmail.com} \\
  \And
  Karl-Günter Technau \\
  Empilweni Services and Research Unit, \\ Department of Paediatrics and Child Health,\\ University of the Witwatersrand, Johannesburg
  \\
  South Africa \\
  \texttt{Karl-Gunter.Technau@wits.ac.za} \\
  \And
    Thokozile Malaba \\
  Division of Epidemiology and Biostatistics,\\ School of Public Health, Faculty of Health Sciences,\\ University of Cape Town,  South Africa \\
  \texttt{thoko.malaba@uct.ac.za} \\
  \And
  Alex Welte \\
  Discipline of Public Health,\\
  School of Medicine,\\
  University of KwaZulu-Natal \\
  South Africa \\
  \texttt{alexwelte@gmail.com} \\
  \And
  Ushma Mehta \\
  Division of Clinical Pharmacology, \\
  Department of Medicine, \\
  University of Cape Town \\
  South Africa \\
  \texttt{ushma.mehta@uct.ac.za} \\
  \And
  Bruce A. Bassett \\
  WITS MIND Institute and CSAM, University of the Witwatersrand, \\
  University of Cape Town \& Grai Labs \\
  Cape Town, South Africa \\
  \texttt{bruce.a.bassett@gmail.com} 
  \\
}

\begin{document}
\maketitle
\footnotetext{$^\star$ Equal contributions.}
\vspace{-0.5cm}
\begin{abstract}
Manual digitisation of structured handwritten documents is slow and costly. We benchmark 17 leading frontier multi-modal large language models and open-source  models against a very challenging real-world medical form that mixes dates; structured, printed text; hand-written responses and significant variability challenges. None of the smaller or older models perform well but the latest Google and OpenAI models reach accuracies around $85\%$ with weighted F1 scores  $\simeq 90\%$ across the discrete or predefined fields despite the very challenging nature of the responses.  
Clear task specific strengths emerge: GPT 5.4 excels in noisy date extraction as well as reliability with the lowest hallucination rate (6\%). Claude Sonnet 4.6 had the best average performance across formatted fields (dates and numerical values), while Gemini 3.1 delivered the best overall performance, with the lowest free text error rates (WER = 0.50 and CER = 0.31) and the strongest results across discrete classification metrics. 
We further show that prompt optimisation dramatically improves macro precision, recall and F1 by over $60\%$, but has little impact on weighted metrics (only $\sim2-5\%$ improvement).  These results provide evidence that the rapid improvements of multimodal large language models offer a compelling pathway toward fully automated digitisation of complex handwritten workflows that is particularly relevant in low- and middle-income countries.

\end{abstract}


\section{Introduction}
The digital transformation of paper-based archives remains one of the most significant challenges in information science. Optical Character Recognition (OCR) has served as the foundational technology for this task for decades; however, its efficacy is often limited by the quality of the source material. While printed, high-contrast documents are now transcribed with near-perfect accuracy, the recognition of unconstrained handwriting characterized by inconsistent stroke widths, overlapping characters, and non-standard symbols continues to push the limits of traditional computer vision.

This challenge is nowhere more apparent than in the healthcare sector. The digitization of medical documentation represents one of the most critical yet persistent bottlenecks in the modernization of healthcare infrastructure. Traditionally, the conversion of handwritten clinical records into structured digital formats has remained a manual, labor-intensive process characterized by high operational costs and significant time latency. This process introduces substantial risks to patient safety through errors, data fragmentation, and the inherent variability of human handwriting.

The emergence of Multimodal Large Language Models (MLLMs) and specialized Vision-Language Transformers offers a potential shift. Unlike traditional OCR, systems that focus solely on character-to-pixel mapping, modern architectures integrate visual perception with deep linguistic reasoning. This study benchmarks these frontier models against open-source alternatives to evaluate their efficacy in digitizing complex, handwritten medical forms within a critical public health context. This study evaluates the efficacy of these frontier models within a high-stakes clinical context, specifically addressing the digitisation requirements of the UBOMI BUHLE project\footnote{\url{https://ubomibuhle.org.za/}}  in South Africa. Using this real-world use case as a benchmark, we aim to determine how advanced AI can solve the specific structural and legibility challenges inherent in complex, handwritten medical forms.

\subsection{Background: The UBOMI BUHLE Project}
The UBOMI BUHLE project's Pregnancy Exposure Registry (UBPER) in South Africa systematically monitors maternal exposure to medications and vaccines to evaluate longitudinal birth outcomes. While its primary focus resides on the safety of HIV/AIDS therapeutics and prophylaxis, the registry’s framework is designed for the clinical assessment of other comorbid conditions, including tuberculosis (TB), obesity, hypertension, and syphilis.

Approximately 15 000 pregnant women are enrolled from 17 antenatal care (ANC) clinics across three South African provinces (KwaZulu Natal, Gauteng and Western Cape). These pregnancies are followed from the first antenatal visit until the time of delivery. Information on the mother’s health, her obstetric history, the timing of her pregnancy, medication taken during pregnancy and the outcome of the pregnancy (e.g. whether the baby was liveborn or stillborn, low birth weight, premature, had any congenital disorders identified at birth) is collected from the patient held maternity case record (MCR) and supplemented with additional data from facility-based registers, the national laboratory service as well as facility-held records. 

Specific details from the MCR are captured by data capturers embedded at sites into a relational database. Registry staff conduct trainings and other efforts at sites to improve clinical care and the quality and completeness of data captured in the MCR which is the primary health record of a pregnant woman during her pregnancy. However these procedures are hampered by the inefficiencies of manual data capture, retrospective data cleaning and delayed feedback to clinicians.

The sentinel sites are overseen by 5 separate project partners each with different data capture systems. Despite this, a standardized data transfer format enables data pooling into a national SQL database, ensuring consistency, pseudonymisation, and security. Pooled data is analysed to assess associations between medical conditions, medication and the risk of an adverse pregnancy outcome (e.g. stillbirth, low birth weight, congenital disorders and preterm birth).

This paper benchmarks the use of multiple frontier and open-source models in their ability to digitise handwritten digital records with the aim of digitising a large number of medical documents into a single database allowing for assessing trends and ease of data access. We evaluate the performance across structured checkboxes, numerical entries, dates and free text fields across a complex set of real maternity forms. 

\subsection{Related Work}

Document Image Analysis (DIA) has evolved from mechanical character recognition to structured semantic interpretation, integrating layout analysis with contextual text extraction to reconstruct meaning from heterogeneous, low-quality materials. Conventional OCR systems often struggle with the irregular and highly variable nature of clinical handwriting \cite{crosilla2025benchmarkinglargelanguagemodels}, particularly under high cognitive load \cite{kumar_interpreting_doctor_notes}. Their modular approach typically results in cumulative error propagation, where minor segmentation inaccuracies cascade into failures in semantic extraction \cite{Lopresti_ocr_errors}.

Prior to the emergence of frontier LLMs, specialized Transformer-based Vision-Language models like TrOCR \cite{li2022trocrtransformerbasedopticalcharacter} and Donut \cite{2021arXiv211115664K} introduced end-to-end architectures that treated document images as sequence-to-sequence problems. While models such as Pix2Struct \cite{lee2023pix2structscreenshotparsingpretraining} and LayoutLM \cite{layoutLM} successfully captured spatial hierarchies, they often required extensive task-specific fine-tuning and struggled with the semantic variability found in clinical notes.

The current paradigm shift toward Multimodal Large Language Models (MLLMs), such as GPT-4o and the Qwen2.5-VL series \cite{bai2025qwen25vltechnicalreport}, integrates general-purpose visual perception with deep linguistic reasoning. These models perform competitive OCR and table extraction \cite{shi2023exploringocrcapabilitiesgpt4vision} by leveraging semantic context to disambiguate illegible characters that would otherwise fail in traditional or specialized transformer pipelines.

\section{Dataset}
\subsection{Dataset Composition}
For initial benchmarking, a synthetic dataset of four forms containing simulated data was used, which did not require POPIA compliance. This dataset allowed testing a wide range of LLMs, including open-source models, to identify the most suitable models for processing the real dataset. The real dataset consisted of 49 documents, with all personal and clinic information fully redacted. Both datasets share the same structure and can be categorized into three types of fields: discrete selection fields, which include checkboxes or predefined lists of possible outcomes; format-constrained fields, which follow an implicit structure such as dates or numerical values; and free-text fields.
\begin{figure}[H]
    \centering
    \includegraphics[width=1\linewidth]{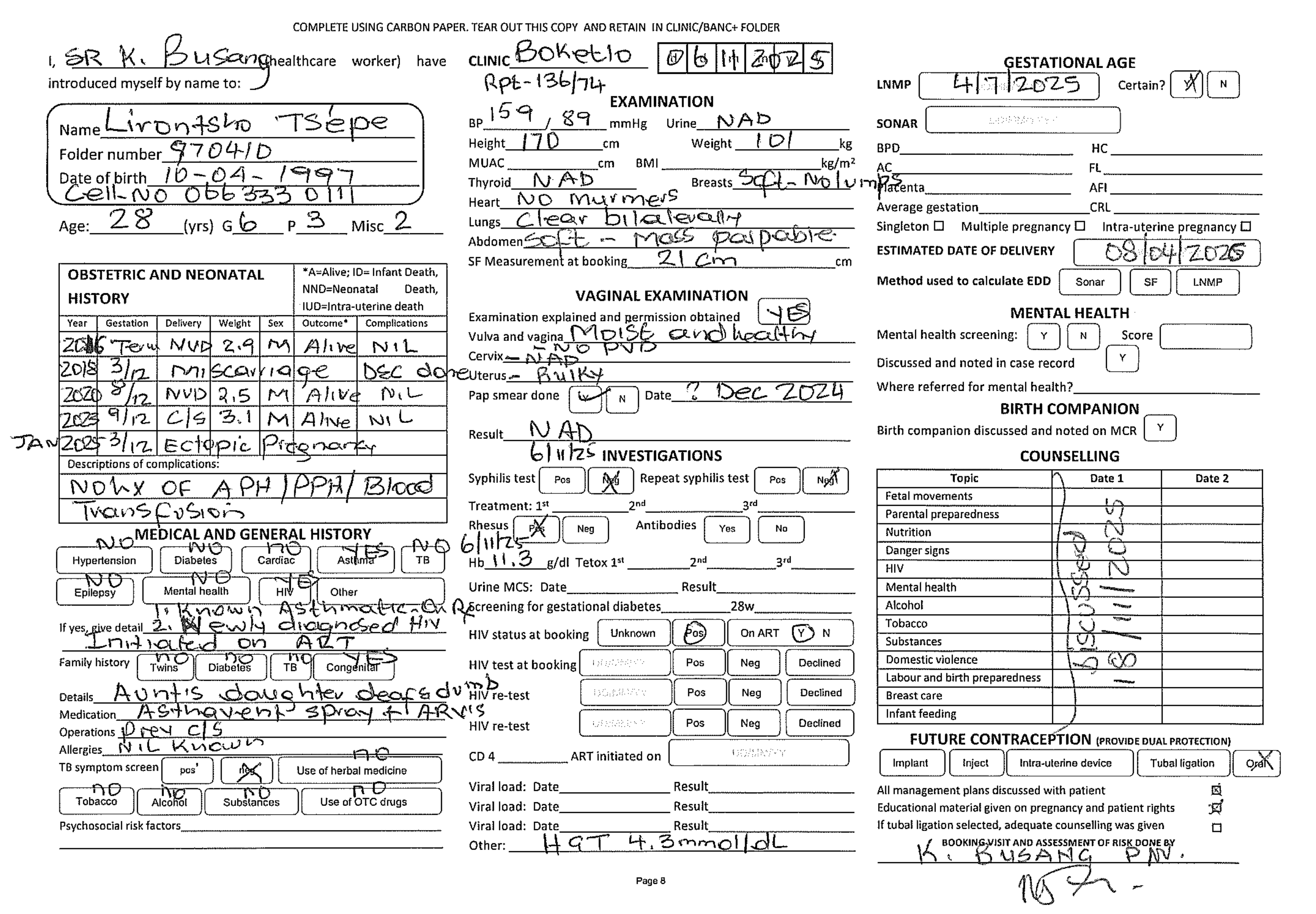}
    \caption{Our maternity case record with example realistic synthetic data. The form shows a challenging mix of dates, abbreviations, idiosyncratic responses and structural anomalies such as text written outside the intended field space. Note the handwriting in the synthetic data is often substantially neater than that of the real data set.} \label{synthetic_data_image}
\end{figure}
The form shown in  Figure~\ref{synthetic_data_image} depicts the various sections and fields which will be present across all forms. The majority of the fields are contained within the following five sections: Obstetric \& Neonatal History, Medical \& General History, Examinations, Vaginal Examinations and Investigations. Gestational Age was often blank with only five of the forms within the real dataset containing the full set of information. The remainder of the sections contain a substantially smaller subset of fields making them more prone to sparsity and small sample size effects.

\subsection{Handwriting and structural challenges/OCR Challenges}
\label{sec:Handwriting_and_structural_challenges/OCR_Challenges}
Transcribing handwritten clinical documents using real-time or near real-time OCR systems presents different challenges. Firstly, the form may have been completed by numerous members of staff resulting in multiple handwriting styles as well as varying levels of training, diligence and familiarity with the documentation and its usual standards.
Secondly, this form is often the minimum documentation required rather than a reflection of the entire patient's journey. For example, there are only three places provided for repeat HIV tests while recent guidelines mandate monthly testing which may include up to eight tests. Meticulous members of staff often squeeze additional information outside of existing spaces or the intended structure of the form.
Finally, clinical notes in the often busy and under-resourced healthcare setting often have characteristics that further complicate automated transcription. The handwriting is often illegible due to intense time pressure and text is often written in a haphazard manner which is most prominent with dates. There are two examples of this present in Figure~\ref{synthetic_data_image}; under the Medical \& General History section in which the condition checkboxes are written "No" instead of being left blank and in the Counselling section the date is written vertically rather than individually for each field. The time pressures described result in a huge number of medical shorthands; for instance, "NAD" is often used to denote "No abnormalities detected", "?" is used to denote "unknown" and the symbol “°” is shorthand for “no”.
This is further complicated by the fact that the clinical staff apply their own encoding conventions when completing forms, replacing categorical responses with codes not defined on the form itself. For instance for various condition fields within the Medical \& General History section "1" is used to denote user, "2" denotes former user and "3" denotes non user.
To compensate for this, a large number of rules and text normalisations are required within the prompt instructing the model how to approach each form and the dataset.  
Most models will need additional prompting to be able to interpret and detect this which would not be required by humans transcribing the information. Though this text normalisation can be done post processing, an automatic agent capable of these text normalisation conditions would be substantially more time and cost effective when dealing with a large data set. Thus, for the purposes of this paper all normalisation conventions are imposed on the model at prompt-level.
\section{Methodology}
\subsection{Image Processing}
Modern VLMs and MLLMs fundamentally depart from the canonical OCR pipeline by adopting an end-to-end-architecture that eliminates the need for discrete, sequential processing stages \cite{2021arXiv211115664K}. Instead of decomposing the recognition task into explicit substeps like segmentation, feature extraction and character classification, the models treat document understanding as a unified sequence-to-sequence problem, whereby the raw image input is directly mapped to a structured textual output \cite{lee2023pix2structscreenshotparsingpretraining}. 

The image processing pipeline in these architectures starts with a Vision Encoder which divides the input image into a grid of fixed size patches. Each patch is then linearly projected into a high-dimensional embedding, which produces a sequence of visual tokens that collectively represent the spatial and semantic content of the entire image. The encoding process preserves structural relationships across the document without the requirement of explicit bounding box detection or character segmentation \cite{li2022trocrtransformerbasedopticalcharacter}. These visual token sequences are projected into the language model's embedding space via a cross-modal alignment layer allowing the language decoder to attend jointly over both visual and textual representations \cite{bai2025qwen25vltechnicalreport}.

The language decoder proceeds to autoregressively generate output text, conditioning each token on both the encoded visual context and the previously generated tokens \cite{bai2025qwen25vltechnicalreport}. Crucially, this generative stage does not perform isolated character classification; instead, it leverages the model’s pre-trained linguistic knowledge to disambiguate visually ambiguous content \cite{crosilla2025benchmarkinglargelanguagemodels}. In the context of clinical document handwriting, this means that the model can resolve illegible characters by inferring the most contextually plausible term from the surrounding information (on the document) \cite{kumar_interpreting_doctor_notes}.
For frontier MLLMs, postprocessing is a minimal step by design because semantic coherence is embedded in the autoregressive generation process and thus the need for separate error correction or contextual refinement stages is significantly reduced \cite{2021arXiv211115664K}. The output is directly serialized into the required format, for the purposes of this study, a structured JSON schema. Open-source pipelines, however, may require additional postprocessing steps such as an artefact removal and spatial metadata stripping, as discussed in Section~\ref{sec:Open_Source_Pipelines}.

\subsection{Frontier LLM Pipeline}
\label{sec:Frontier_LLM_Pipeline}
Each model was given two opportunities to output a valid output in a JSON format. The temperature was set to zero to ensure that the models are as close to deterministic as possible.

Each medical form image is encoded and transmitted alongside a carefully curated system prompt to the respective model. This prompt is constructed from a plain-text instruction file and a JSON output template, providing the model with both the extraction directive and required output schema. The model subsequently performs visual perception and structured extraction in a single forward pass \cite{achiam2023gpt4,bai2025qwen25vltechnicalreport} with temperature set to zero to ensure a deterministic output across repeated evaluations.

Should the call to the Application Programming Interface (API) fail, the pipeline employs an exponential backoff before reattempting. Images that cannot be processed after exhausting all tries are logged and skipped, thereby ensuring the remainder of the benchmarking proceeds smoothly. The raw model response is subsequently parsed into a structured JSON object; if strict parsing fails as a result of minor formatting artifacts in the model output, an automated repair strategy is then applied. If parsing cannot be recovered, the file is logged and excluded from evaluation \footnote{When individual fields 
within a successfully parsed form cannot be interpreted, models do not flag 
uncertainty. They will either return a blank value, infer the closest plausible alternative, or hallucinate some output entirely. The hallucination rates across 
the most prevalent fields is reported in Section~\ref{sec:Results_and_Discussion}}.

The evaluation process follows a recursive, field-level comparison between the model prediction and the associated ground truth JSON file. Both values are normalised (case and whitespace) before an exact match is checked. Fields that are present in the ground truth and are missing or incorrectly populated in the prediction are denoted as errors, and fields hallucinated by the model that do not appear in the ground truth are also penalized. To measure accuracy, all section scores of the form are aggregated.

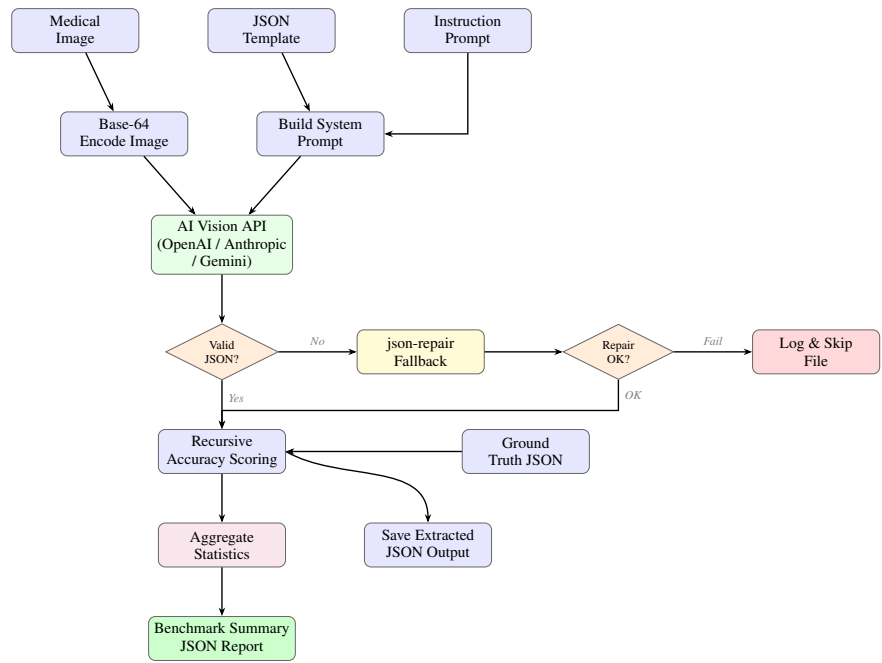
\begin{figure}[H]
\centering
\scalebox{0.65}{%
\begin{tikzpicture}[
    node distance = 0.6cm and 1.4cm,
    box/.style    = {rectangle, rounded corners=4pt, draw=black!60,
                     fill=blue!10, minimum width=2.6cm, minimum height=0.9cm,
                     align=center, font=\small},
    decision/.style={diamond, draw=black!60, fill=orange!15,
                     minimum width=2.2cm, minimum height=0.9cm,
                     align=center, font=\scriptsize, aspect=2},
    arrow/.style  = {-{Stealth[length=5pt]}, thick},
    label style/.style = {font=\scriptsize\itshape, text=gray}
]

\node[box] (img)   {Medical\\Image};
\node[box, right=of img]  (tmpl)  {JSON\\Template};
\node[box, right=of tmpl] (instr) {Instruction\\Prompt};

\node[box, below=1.2cm of img, xshift=1.0cm] (encode) {Base-64\\Encode Image};
\node[box, right=1.4cm of encode] (build)  {Build System\\Prompt};

\node[box, below=1.2cm of encode, xshift=2.0cm, fill=green!10] (api)
    {AI Vision API\\(OpenAI / Anthropic\\/ Gemini)};

\node[decision, below=1.0cm of api] (valid) {Valid\\JSON?};

\node[box, right=1.6cm of valid, fill=yellow!20] (repair) {json-repair\\Fallback};
\node[decision, right=1.6cm of repair] (repaired) {Repair\\OK?};
\node[box, right=1.6cm of repaired, fill=red!15] (skip) {Log \& Skip\\File};

\node[box, below=1.0cm of valid] (score) {Recursive\\Accuracy Scoring};

\node[box, right=1.6cm of score, xshift=2cm] (gt) {Ground\\Truth JSON};

\node[box, below=1.0cm of score, fill=purple!10] (agg)
    {Aggregate\\Statistics};

\node[box, right=1.6cm of agg] (save) {Save Extracted\\JSON Output};

\node[box, below=1.0cm of agg, fill=green!20] (report)
    {Benchmark Summary\\JSON Report};

\draw[arrow] (img)   -- (encode);
\draw[arrow] (tmpl)  -- (build);
\draw[arrow] (instr) |- (build);

\draw[arrow] (encode) -- (api);
\draw[arrow] (build)  -- (api);

\draw[arrow] (api)   -- (valid);

\draw[arrow] (valid) -- node[right, label style, yshift=4pt]{Yes} (score);
\draw[arrow] (valid) -- node[above, label style]{No}  (repair);

\draw[arrow] (repair)   -- (repaired);
\draw[arrow] (repaired) -- node[above, label style]{Fail} (skip);
\draw[arrow] (repaired) -- node[right, label style]{OK} ++(0,-1.2cm) -| (score);

\draw[arrow] (gt) -- (score);

\draw[arrow] (score) -- (agg);
\draw[arrow] (score.east) to[out=-30,in=90] (save.north);

\draw[arrow] (agg) -- (report);

\end{tikzpicture}%
}
\caption{End-to-end pipeline used for testing frontier LLM's on our medical benchmark dataset.}
\label{fig:pipeline}
\end{figure}

\subsection{Open Source Pipelines}
\label{sec:Open_Source_Pipelines}

To evaluate the digitization of handwritten medical documents, we consider the following multimodal models: dots.ocr~\cite{li2025dotsocrmultilingualdocumentlayout}, Qwen2-VL-2B \cite{wang2024qwen2vl}, and Qwen2.5-VL-3B \cite{bai2025qwen25vltechnicalreport}. For dots.ocr, a two-stage decoupled extraction pipeline was developed that separates the visual task of reading handwriting from the linguistic task of structuring the data. For the Qwen models, we instead use a single stage architecture that combines both the reading of handwriting and the parsing.

The difference in pipeline structure occurs with dots.ocr. Here, the process is broken down into two distinct steps. The first of these steps involves visual perception and raw text extraction where the vision encoder processes the standardised image to extract all readable text. Models are loaded with 4-bit quantization and bfloat16 precision for computational efficiency. Each model receives the image alongside a guiding prompt and relies on its native vision-language architecture to map visual patches to text tokens. A post-processing function then strips any spatial metadata or residual HTML artefacts. The second step involves linguistic parsing and information structuring, where a dedicated language model (Qwen2.5-7B-Instruct \cite{bai2025qwen25vltechnicalreport}) receives the raw OCR output alongside JSON schema and the prompt, with domain-specific normalization rules. These rules address abbreviations, date, gestation formatting, value cleaning, among others.

The Qwen2-VL-2B and Qwen2.5-VL-3B single-stage pipelines are largely identical to the frontier LLM pipeline, with the exception that the models are run locally as opposed to an external API call. Metric evaluation for all opensource models follows the same recursive, field-level comparison procedure described in Section~\ref{sec:Frontier_LLM_Pipeline}.
\subsection{Prompt Engineering}
Recent advances in vision-language models have enabled the integration of handwritten text recognition with structured reasoning, allowing for schema-constrained extraction, domain-specific rule enforcement, etc. at inference time. Unlike fine-tuned models, frontier LLMs are very sensitive to phrasing, specificity, and structure of the input prompt, thus making prompt engineering a crucial means for improving digitisation accuracy.

For the purposes of this study two distinct prompts and templates were used. The prompt contained general and normalisation rules while the template was entirely blank for the base template (excluding field names) and the optimal template contained discretisation of various discrete selection fields. For example, delivery type as ["NVD", "C/S", "TOP"...], and HIV status as ["Un-known", "Positive", "Negative"] which directly constrains the models output space at the schema level as opposed to narrative instruction. Beyond the discretisation, the optimised prompt encoded 16 domain-level specific normalisation rules covering date formatting (DD/MM/YYYY with zero-padding and year expansion), abbreviation resolution, checkbox encoding conventions, blood pressure field splitting, gestation during duration formatting, and clinical shorthand interpretation, as discussed in Section~\ref{sec:Handwriting_and_structural_challenges/OCR_Challenges}. These rules specifically address transcription behaviours that could contribute to higher post-processing costs or produce inconsistent outputs across the various models. The extent to which these optimisations contribute to improved extraction accuracy, structural consistency, and reduced hallucinations is evaluated in Section~\ref{sec:Results_and_Discussion}.

\subsection{Evaluation Metrics}
The primary evaluation metric is the \textbf{strict string global macro average accuracy}, which assesses exact matches between the model-generated string and the ground truth for each field. To account for the varying nature of the data, fields were categorised as follows:
\begin{itemize}
    \item \textbf{Discrete Selection Fields:} Evaluated using precision, recall, and F1-score within a multi-class classification framework, where "null" values were treated as a distinct class.
    \item \textbf{Formatted fields:} Evaluated based on adherence to an expected output format without a predefined discrete list of options. Examples include dates which must adhere to the structure DD/MM/YYYY and numerical values (often accompanied with units).
    \item \textbf{Free-text fields:} Evaluated using Word Error Rate (WER) and Character Error Rate (CER).
\end{itemize}

For global, formatted, and sectional accuracy, we report the median along with the 16th and 84th percentiles of the observed empirical distribution to characterise the variation in performance from form to form.

\section{Results and Discussion}
\label{sec:Results_and_Discussion}
\subsection{Performance on Synthetic Data}

We explored the following 17 models across a few families: Qwen (2, 2.5) \cite{wang2024qwen2vl, bai2025qwen25vltechnicalreport}, Dots.ocr \cite{li2025dotsocrmultilingualdocumentlayout}, Claude (Haiku 4.5, Opus 4.6, Sonnet 4.6) \cite{anthropic2025haiku, anthropic2026opus, anthropic2025sonnet}, Grok (4.1-fast-reasoning) \cite{xai2025grok41}, OpenAI (GPT-4.1, gpt-5-mini
, GPT 5-Nano, GPT-5.2, GPT 5.2-Reasoning (Pro), GPT-5.4) \cite{achiam2023gpt4, openai2025gpt41, openai2025gpt5}
and Google Gemini (2.5 Flash, 2.5 Pro, 3.1 Flash Lite, 3.1 Pro) \cite{geminiteam2025gemini25, google2025gemini31flashlite, google2025gemini31pro}
The newer and larger models drastically outperformed  smaller and older models as evident in Table ~\ref{tab:small_models}. Opensource models were relatively ineffective scoring the lowest accuracy across filled fields (excluding extremely budget options such as GPT Nano and Mini). 
\begin{figure}[H]
    \centering
    \includegraphics[width=0.59\linewidth]{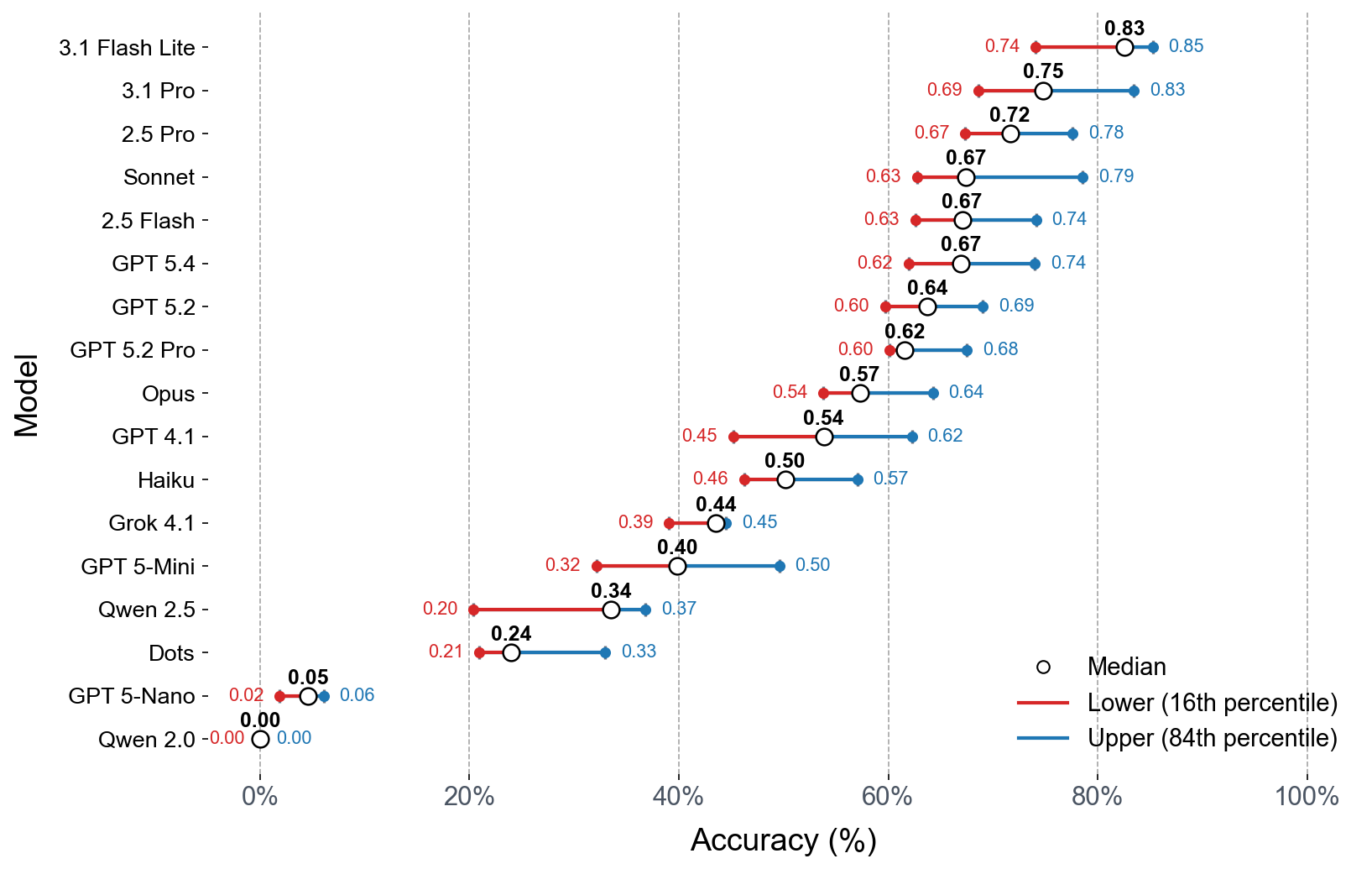}
    \caption{Forest plot displaying the median, 16th and 84th percentile for the overall accuracy excluding null fields across the synthetic data set on the optimal prompt. This is indicative that the small, open-source models were the weakest performing by a large margin $\geq 40\%$ from best performing models.}
    \label{fig:synthetic forest}
\end{figure}
As previously mentioned, the synthetic data set consists of four simulated forms which were used to determine high-level patterns within model performance. The median (alongside the 16th and 84th percentiles) as evident in figure ~\ref{fig:synthetic forest} provide an overview of overall model performance. Newer and larger models consistently outperformed their smaller and older counterparts. Frontier models such as Gemini 3.1 Flash Lite achieved 86\% overall accuracy and 83\% when excluding empty fields, closely followed by Gemini 3.1 Pro (75\% excluding empty fields). In contrast, Open-source models (Qwen and Dots) were substantially less effective. Qwen 2.5 was the best performing Open-source model achieving only 34\% overall accuracy when excluding empty fields. 

\begin{table}[H]
\centering
\scriptsize
\caption{Performance metrics of various LLMs on the synthetic dataset using the optimized prompt and template. Metrics show median values with 16th and 84th percentiles; WER is word error rate CER is character error rate---WER and CER were only calculated across free text fields. Hallucination rate indicates proportion of fabricated outputs. The open-source models are shown in the last three rows and are not competitive in every metric. Gemini 3.1 Flash Lite followed by 3.1 Pro were the best performing models across the synthetic data set. Haiku 4.5 and GPT 5-Mini were unable to interpret the discrete option lists, often returning the list verbatim, which drastically decreased performance.}
\label{tab:small_models}
\resizebox{0.6\textwidth}{!}{
\renewcommand{\arraystretch}{1.6}
\begin{tabular}{|l| c c c| c c| c c|}
\hline
\shortstack{Model} & \shortstack{Accuracy} & \shortstack{Accuracy\\(w/o Nulls)} & \shortstack{Formatted\\Accuracy} & \shortstack{WER} & \shortstack{CER} & \shortstack{Null Field\\Accuracy} & \shortstack{Hallucination\\Rate} \\
\hline
3.1 Pro      & $0.84_{-0.08}^{+0.07}$ & $0.75_{-0.06}^{+0.09}$ & $0.84_{-0.18}^{+0.12}$ & \textbf{0.33} & 0.15 & \textbf{0.93} & \textbf{0.07} \\
3.1 Flash Lite   & \bm{$0.86$}$_{-0.09}^{+0.05}$ & \bm{$0.83$}$_{-0.08}^{+0.03}$ & \bm{$0.91$}$_{-0.17}^{+0.04}$ & 0.40 & 0.21 & 0.91 & 0.09 \\
2.5 Pro      & $0.79_{-0.07}^{+0.06}$ & $0.72_{-0.04}^{+0.06}$ & $0.76_{-0.12}^{+0.10}$ & 0.39 & 0.18 & 0.87 & 0.13 \\
2.5 Flash   & $0.75_{-0.04}^{+0.07}$ & $0.67_{-0.04}^{+0.07}$ & $0.71_{-0.12}^{+0.14}$ & 0.45 & 0.26 & 0.86 & 0.14 \\
GPT 5.4     & $0.80_{-0.08}^{+0.05}$ & $0.67_{-0.05}^{+0.07}$ & $0.73_{-0.13}^{+0.12}$ & 0.47 & 0.21 & 0.91 & 0.09 \\
GPT 5.2-Pro  & $0.76_{-0.08}^{+0.07}$ & $0.62_{-0.01}^{+0.06}$ & $0.72_{-0.14}^{+0.12}$ & 0.60 & 0.38 & 0.89 & 0.11 \\
GPT 5.2     & $0.76_{-0.08}^{+0.07}$ & $0.64_{-0.04}^{+0.05}$ & $0.69_{-0.13}^{+0.14}$ & 0.51 & 0.23 & 0.88 & 0.12 \\
GPT 5-Mini  & $0.60_{-0.12}^{+0.08}$ & $0.40_{-0.08}^{+0.10}$ & $0.76_{-0.16}^{+0.14}$ & 0.60 & 0.33 & 0.77 & 0.23 \\
GPT 4.1      & $0.67_{-0.10}^{+0.10}$ & $0.54_{-0.09}^{+0.08}$ & $0.69_{-0.19}^{+0.16}$ & 0.73 & 0.46 & 0.82 & 0.18 \\
GPT 5-Nano        & $0.46_{-0.08}^{+0.04}$ & $0.05_{-0.03}^{+0.01}$ & $0.64_{-0.15}^{+0.13}$ & 0.98 & 0.93 & 0.89 & 0.11 \\
Grok 4.1     & $0.67_{-0.10}^{+0.04}$ & $0.44_{-0.05}^{+0.01}$ & $0.67_{-0.11}^{+0.09}$ & 0.82 & 0.66 & 0.90 & 0.10 \\
Haiku 4.5       & $0.62_{-0.08}^{+0.07}$ & $0.50_{-0.04}^{+0.07}$ & $0.58_{-0.11}^{+0.21}$ & 0.71 & 0.50 & 0.74 & 0.26 \\
Sonnet 4.6     & $0.77_{-0.07}^{+0.09}$ & $0.67_{-0.05}^{+0.11}$ & $0.69_{-0.12}^{+0.18}$ & 0.36 & \textbf{0.14} & 0.87 & 0.13 \\
Opus 4.6       & $0.73_{-0.06}^{+0.06}$ & $0.57_{-0.03}^{+0.07}$ & $0.74_{-0.17}^{+0.11}$ & 0.64 & 0.33 & 0.88 & 0.12 \\
\hline
Qwen 2.5     & $0.48_{-0.13}^{+0.07}$ & $0.34_{-0.13}^{+0.03}$ & $0.60_{-0.27}^{+0.13}$ & 0.81 & 0.67 & 0.64 & 0.36 \\
Qwen 2       & $0.43_{-0.09}^{+0.04}$ & $0.00_{-0.00}^{+0.00}$ & $0.61_{-0.18}^{+0.12}$ & 1.00 & 1.00 & 0.87 & 0.13 \\
Dots        & $0.39_{-0.02}^{+0.06}$ & $0.24_{-0.03}^{+0.09}$ & $0.24_{-0.01}^{+0.04}$ & 1.34 & 1.31 & 0.59 & 0.41 \\
\hline
\end{tabular}}
\renewcommand{\arraystretch}{1.0}
\begin{minipage}{\textwidth}
\smallskip
Word Error Rate (WER) and Character Error Rate (CER) measure the proportion of words and characters respectively that must be inserted, deleted, or substituted to match the ground truth.
\end{minipage}
\end{table}

However, performance does not seem to strongly correlate with reasoning ability. Gemini Flash Models were consistently able to perform better or comparable to their Pro counterparts despite being provided with a substantially lower quantity of reasoning tokens. This is also apparent within the GPT family, GPT 5.4 was able to outperform the substantially more expensive 5.2 Pro especially in CER and WER. This is indicative that in some instances newer models may have improved performance compared with older, larger models despite a large cost or computational differential. The Gemini model family was extremely dominant within the data set with newest 3.1 models performing best in all metrics (excluding CER in which 3.1 Pro performed marginally worse than Sonnet 4.6). There was an apparent minimum reasoning capability required as Open-source and budget models such as GPT 5-Mini and GPT 5-Nano were often unable to interpret the discretised template, producing the list verbatim rather than selecting the correct option. These models performed better with minimal information subsequently requiring less reasoning capabilities as evident within the ablated real dataset \ref{Ablation}.

Formatted fields, as previously mentioned, refers to fields with an implicit structure---dates or numerical values. Models were often asked to normalise to a specific structure, for instance all dates were required in the format DD/MM/YYYY and certain units may have been required for consistency across the data set---testing inference, normalisation capabilities and the precision when following instructions. Gemini 3.1 Flash Lite followed by 3.1 Pro were once more the best performing models within this category. 

For the evaluation of free text fields we used the Character Error Rate (CER) and Word Error Rate (WER). The WER across all models were relatively high with 3.1 Pro achieving the lowest WER (0.33). This indicates that all models are not able to reliably extract all words within a given free text field and often exclude or add phrases. This is likely due to free text fields being sparse within the forms, being the most difficult fields, often containing long poorly transcribed text and the unique medical abbreviations which were often excluded by models (for instance "PXX" and "RXX" for pulse rate and respiratory rate respectively). The CER was substantially lower across the majority of models with Sonnet 4.6 being the best performer (0.14) indicating that fewer character-level errors are present when the words are present. 

For null field accuracy, most models performed well, having correctly identified empty fields most of the time($\geq85\%$). Gemini 3.1 Pro again led with 93\% accuracy, though hallucination rates remain non-negligible across all models, the lowest observed at 7\% for the same model which can pose major concerns for sensitive and medical data sets.  

\subsection{Performance on Real Data}
Model performance was not the only criteria considered when choosing which models to use for the real data set. GPT 5.2 Pro was excluded due to long processing times and cost that did not correlate with performance, making it ineffective for digitisation of large data sets. GPT 4.1 was excluded as it performs substantially weaker than its newer GPT counterparts and may soon be deprecated. Finally, Open-source models (Qwen and Dots) as well as GPT 5-Nano were excluded due to their weak performance. The final model selection is as follows: Claude (Haiku 4.5, Opus 4.6, Sonnet 4.6), OpenAI (GPT 5-Mini, GPT 5.4) and Google Gemini (2.5 Flash, 2.5 Pro, 3.1 Flash Lite, 3.1 Pro). 

These models were tested on the entire real patient data set consisting of 49 forms. The Claude model family were unable to process three forms within the data set due to image size constraints which were excluded from its evaluation. Additionally Gemini 3.1 Pro produced two truncated outputs, most likely due to the model entering infinite reasoning loops. These truncated outputs were included in the data set and assigned 0\% accuracy for missing fields. Gemini 3.1 Pro was also prone to structural errors, often choosing arrays when lists were required within the template format. However, models were not penalised for template based structural errors as the JSON structure was flattened prior to evaluation. 
\begin{figure}[H]
    \centering
    \includegraphics[width=0.56\linewidth]{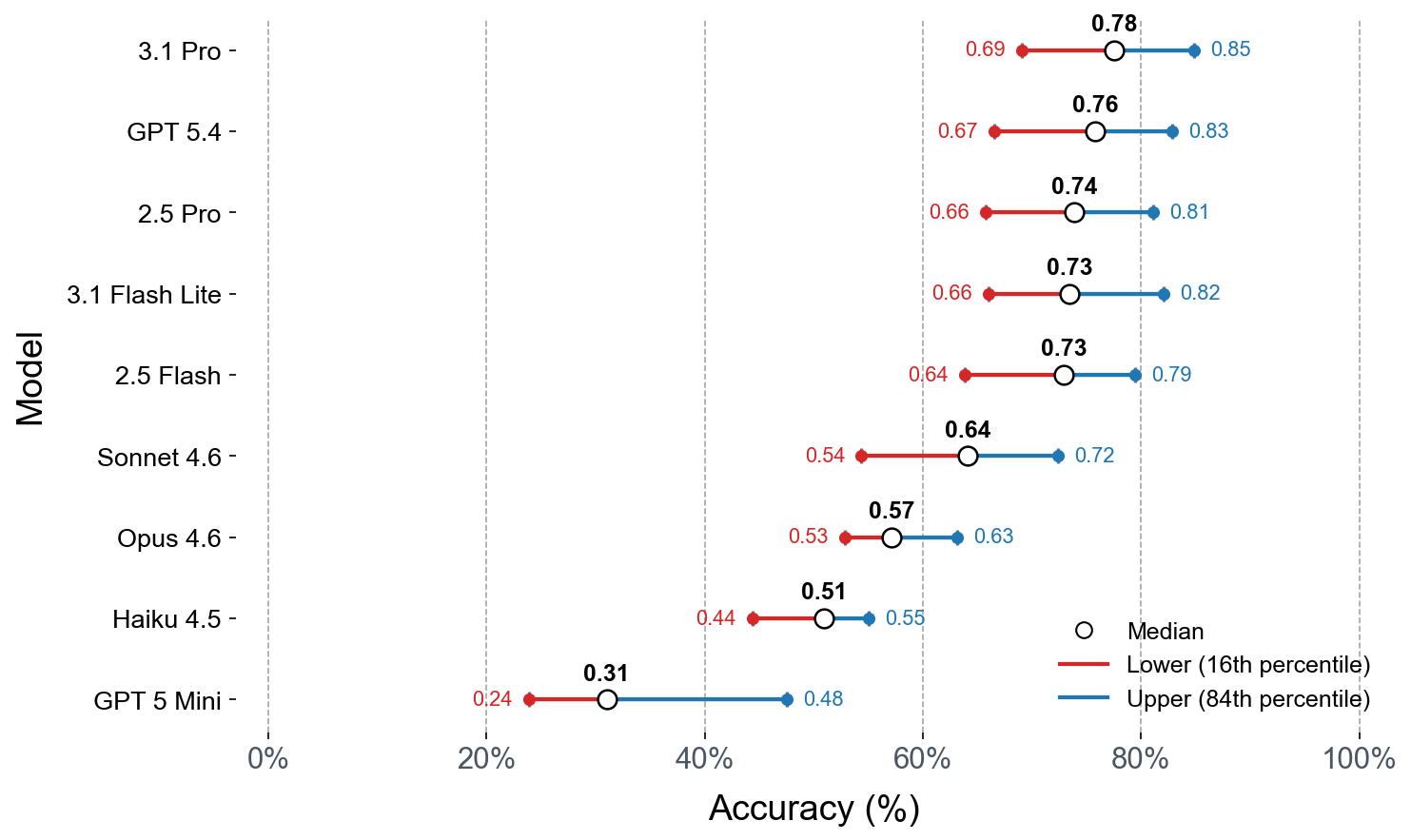}
    \caption{Forest plot displaying the median, 16th and 84th percentile for the overall accuracy excluding null fields across the real data set on the optimal prompt. Frontier Gemini and GPT models performed best with very little performance deviation across these models.}
    \label{forest}
\end{figure}
Figure ~\ref{forest} provides the overall performance excluding empty fields across the real-world data set. Transitioning from the synthetic data set to the real data set resulted in minor differences. The deviation between different models was substantially smaller within the majority of frontier models performing relatively similar (within 5\%). The Claude family was an exception to this as they proved to perform substantially weaker than their Gemini and GPT counterparts ($\sim$10\%+). Gemini 3.1 Flash suffered a large drop in median performance from 83\% in synthetic data to 73\% in the real data, likely an artifact of the small synthetic data sample size misrepresenting its performance capabilities. Gemini 3.1 Pro (78\%) and GPT 5.4 (76\%) achieved the highest median accuracies followed closely by the remainder of the Gemini family. As within the synthetic data set, GPT 5-Mini was unable to interpret the discretised template resulting in exceptionally poor performance on the optimal prompt. Model performance was relatively stable with the majority of models being slightly negatively skewed (evident through the 16th and 84th percentiles). 

\begin{table}[H]
\centering
\scriptsize
\caption{Performance metrics of various LLMs on the real-world document dataset using the optimized prompt and template. Metrics include median values with 16th and 84th percentiles; WER is word error rate, CER is character error rate---WER and CER were only calculated for free text fields. Hallucination rate indicates proportion of fabricated outputs. Gemini 3.1 Pro had the best performance overall and across free text fields minimising the CER and WER. GPT 5.4 had the lowest hallucination followed closely by the Gemini 3.1 family. Sonnet 4.6 had the best performance across formatted fields (dates and numerical values).}
\label{tab:metrics_merged}
\resizebox{0.8\textwidth}{!}{
\renewcommand{\arraystretch}{1.6}
\begin{tabular}{|l| r| r r r r| r r| r r|}
\hline
\shortstack{Model} &
\shortstack{Docs} &
\shortstack{Accuracy} &
\shortstack{Accuracy\\(w/o nulls)} &
\shortstack{Formatted\\Accuracy} &
\shortstack{Formatted\\Accuracy\\(w/o nulls)} &
\shortstack{WER} &
\shortstack{CER} &
\shortstack{Null Field\\Accuracy} &
\shortstack{Hallucination\\Rate} \\
\hline
3.1 Pro       & 49 & \bm{$0.85$}$_{-0.05}^{+0.05}$ & \bm{$0.78$}$_{-0.09}^{+0.07}$ & $0.88_{-0.16}^{+0.06}$ & $0.67_{-0.25}^{+0.12}$ & \textbf{0.50} & \textbf{0.31} & 0.92 & 0.08 \\
3.1 Flash Lite & 49 & $0.84_{-0.05}^{+0.04}$ & $0.73_{-0.07}^{+0.09}$ & $0.85_{-0.07}^{+0.08}$ & $0.72_{-0.10}^{+0.10}$ & 0.53 & 0.40 & 0.92 & 0.08 \\
2.5 Pro       & 49 & $0.79_{-0.05}^{+0.06}$ & $0.74_{-0.08}^{+0.07}$ & $0.85_{-0.07}^{+0.06}$ & $0.27_{-0.08}^{+0.12}$ & 0.60 & 0.40 & 0.86 & 0.14 \\
2.5 Flash     & 49 & $0.80_{-0.04}^{+0.04}$ & $0.73_{-0.09}^{+0.07}$ & $0.82_{-0.10}^{+0.07}$ & $0.32_{-0.06}^{+0.13}$ & 0.60 & 0.45 & 0.89 & 0.11 \\
GPT 5.4              & 49 & \bm{$0.85$}$_{-0.05}^{+0.04}$ & $0.76_{-0.09}^{+0.07}$ & \bm{$0.89$}$_{-0.10}^{+0.07}$ & $0.39_{-0.07}^{+0.13}$ & 0.56 & 0.43 & \textbf{0.94} & \textbf{0.06} \\
GPT 5-Mini           & 49 & $0.57_{-0.06}^{+0.09}$ & $0.31_{-0.07}^{+0.17}$ & $0.74_{-0.10}^{+0.11}$ & $0.72_{-0.13}^{+0.12}$ & 0.72 & 0.64 & 0.83 & 0.17 \\
Haiku 4.5            & 46 & $0.65_{-0.09}^{+0.06}$ & $0.51_{-0.07}^{+0.04}$ & $0.65_{-0.12}^{+0.08}$ & $0.73_{-0.31}^{+0.15}$ & 0.82 & 0.69 & 0.77 & 0.23 \\
Sonnet 4.6           & 46 & $0.75_{-0.05}^{+0.07}$ & $0.64_{-0.10}^{+0.08}$ & $0.74_{-0.08}^{+0.06}$ & \bm{$0.77$}$_{-0.29}^{+0.12}$ & 0.56 & 0.37 & 0.88 & 0.12 \\
Opus 4.6             & 46 & $0.73_{-0.06}^{+0.05}$ & $0.57_{-0.04}^{+0.06}$ & $0.70_{-0.08}^{+0.09}$ & $0.36_{-0.14}^{+0.27}$ & 0.72 & 0.54 & 0.90 & 0.10 \\
\hline
\end{tabular}}
\renewcommand{\arraystretch}{1.0}
\end{table}

Although overall performance across frontier models was approximately equivalent, notable differences emerge when examining secondary metrics (Table~\ref{tab:metrics_merged}). Formatted accuracy including empty fields is high across all models, with GPT-5.4 achieving the highest median accuracy (89\%). However, this metric is heavily inflated by the presence of null fields, which artificially boosts performance due to aforementioned sparsity in medical forms. Consequently, formatted accuracy---excluding null fields---provides a more meaningful measure of true extraction capability.

As expected, performance drops substantially once null fields are removed. Under this more stringent metric, Sonnet 4.6 performs best, achieving a median formatted accuracy of 77\%. Interestingly, several lower cost models---Haiku 4.5 (73\%), GPT 5-Mini (72\%), and Gemini 3.1 Flash Lite (72\%)---closely follow, with Gemini-3.1 Pro (67\%) slightly behind. This suggests that performance on structured formatting tasks is not strongly correlated with model scale or reasoning capability. Furthermore, advanced reasoning may sometimes be detrimental, potentially due to unnecessary normalisation of already well-defined fields.

With respect to the distributional behaviour, the top-performing models exhibit negative skew, indicating more consistent high-end performance with occasional lower outliers, whereas weaker models tend to show positive skew, reflecting less reliable outputs with sporadic strong results. Additionally, there is substantial central interval variability. This volatility is likely due to differences in form complexity, though it may also reflect model-specific strengths in handling particular field types (e.g., numerical values versus dates).

The free text field error metrics, CER and WER, increased considerably when shifting from the synthetic data set to the real data set. This indicates that the free text fields within the real data set are substantially more challenging---likely due to the weaker transcription or combination of additional, unknown factors contributing to the writing. The higher WER is indicative that there may a large number of omissions resulting in partial extractions of the full string as numerous free text fields contain medical abbreviations that are not standard for instance "P83", "APH", "T:37.2$^\circ$" which models frequently omit. Gemini 3.1 Pro achieved the best performance across both of these metrics (WER= 0.50, CER= 0.31). 

Finally, hallucination rates are relatively stable across models, with minimal variation among top performers. GPT-5.4 achieves the lowest hallucination rate (6\%). It is important to note that these relatively low hallucination rates are still harmful in a medical context as the cost of a single mistake is very high. 

\begin{table}[H]
\centering
\tiny
\caption{Performance metrics of discrete selection fields across the real data using the optimised prompt indicating the precision, recall and F1 on a macro and weighted level. Gemini 3.1 Pro outperformed all other models across weighted and macro metrics indicating it is the best model for discrete selection data.}
\label{tab:metrics_optimised}
\fontsize{7.5pt}{10pt}\selectfont
\renewcommand{\arraystretch}{1}
\begin{tabular}{|l| r r r r r r|}
\hline
\shortstack{Model} &
\shortstack{Macro\\Precision} &
\shortstack{Macro\\Recall} &
\shortstack{Macro\\F1} &
\shortstack{Weighted\\Precision} &
\shortstack{Weighted\\Recall} &
\shortstack{Weighted\\F1} \\
\hline
3.1 Pro        & \textbf{0.64} & \textbf{0.70} & \textbf{0.66} & \textbf{0.92} & \textbf{0.91} & \textbf{0.91} \\
3.1 Flash Lite & 0.58 & 0.58 & 0.56 & 0.87 & 0.87 & 0.87 \\
2.5 Pro        & 0.46 & 0.50 & 0.46 & 0.88 & 0.84 & 0.85 \\
2.5 Flash      & 0.52 & 0.57 & 0.51 & 0.89 & 0.86 & 0.87 \\
GPT 5.4        & 0.54 & 0.60 & 0.56 & 0.89 & 0.89 & 0.89 \\
GPT 5-Mini      & 0.22 & 0.13 & 0.15 & 0.66 & 0.31 & 0.40 \\
Haiku 4.5& 0.33 & 0.39 & 0.34 & 0.79 & 0.74 & 0.75 \\
Sonnet 4.6& 0.43 & 0.51 & 0.45 & 0.87 & 0.82 & 0.84 \\
Opus 4.6& 0.46 & 0.49 & 0.46 & 0.85 & 0.83 & 0.83 \\
\hline
\end{tabular}
\renewcommand{\arraystretch}{1.0}
\end{table}

Discrete selection field performance was evaluated using both macro-averaged and weighted precision, recall, and F1 scores (Table~\ref{tab:metrics_optimised}). Across all models, macro-level metrics are consistently low; this is highly likely a direct result of significant class imbalance within the dataset. Many discrete selection fields contain rare subclasses, and because macro metrics assign equal weight to each class, misclassification of these infrequent categories is heavily penalised.

In contrast, weighted metrics are substantially higher across all models, indicating strong performance on the most frequently occurring classes. However, these scores are likely inflated by the dominance of the null class, which appears far more often than other categories and disproportionately influences weighted averages. As such, weighted metrics may overestimate true classification performance in practical settings where minority classes are substantially more valuable (such as in a clinical setting).

Among all models, Gemini 3.1 Pro achieves the strongest performance across both macro and weighted metrics, suggesting superior capability in handling categorical classification tasks within complex document layouts. Nevertheless, the discrepancy between macro and weighted scores highlights a key limitation in reliably identifying rare subclasses.

Expanding the dataset would enable a more robust evaluation of these minority classes and help disentangle whether the observed performance degradation in macro metrics arises primarily from dataset imbalance or from inherent model limitations---such as a tendency to default to the most probable class when uncertainty is high.
\begin{figure}[H]
    \centering
    \includegraphics[width=0.8\linewidth]{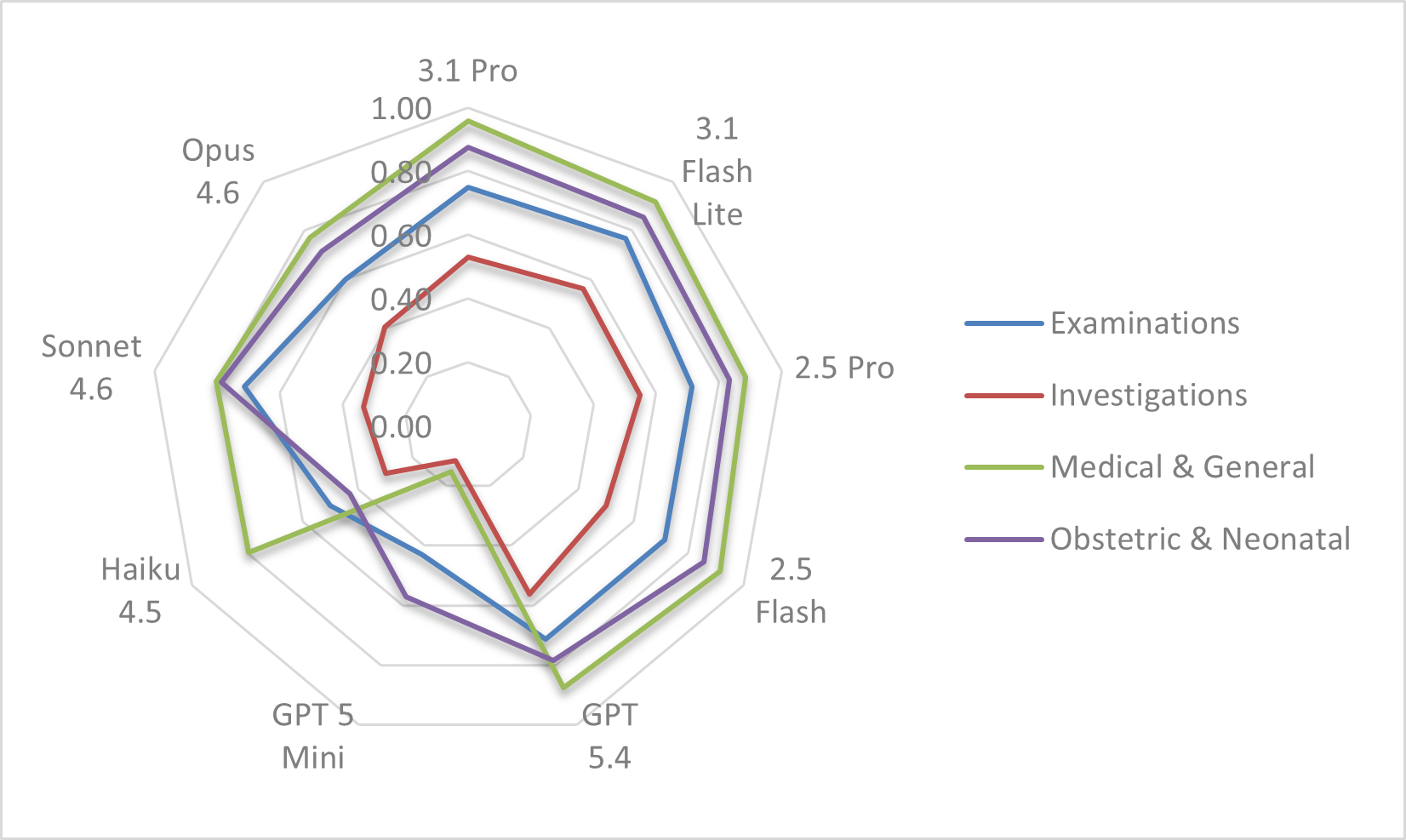}
    \caption{Sectional accuracy across the largest sections in the real data set for each model using the optimal prompt. "Investigations" was the weakest performing section as a result of disordered formatted field transcription, the worst performing free text field and inconsistent normalisation and medical shorthands. "Medical and General" was the best performing field as it primarily consists of discrete selection fields.}
    \label{fig:placeholder}
\end{figure}
\begin{table}[H]
\centering
\caption{Median accuracy scores and performance distribution bounds for each model across specific document sections for real-world data, with null values excluded to measure extraction success on populated fields. "Medical \& General History" and "Future Contraception" are in majority discrete selection fields and achieve the highest performance. Investigations achieves the weakest performance due to disordered transcription.}
\label{tab:median_bounds}

\fontsize{9pt}{10pt}\selectfont

\resizebox{\textwidth}{!}{
\renewcommand{\arraystretch}{1.6}
\begin{tabular}{|l|cccccccccccc|}
\hline
\shortstack{Model} &
\shortstack{Birth\\Companion} &
\shortstack{Booking\\Visit} &
\shortstack{Counselling} &
\shortstack{Examination} &
\shortstack{Future\\Contraception} &
\shortstack{Gestational\\Age} &
\shortstack{Header\\ID} &
\shortstack{Investigations} &
\shortstack{Medical \&\\Gen History} &
\shortstack{Mental\\Health} &
\shortstack{Obstetric \&\\Neonatal History} &
\shortstack{Vaginal\\Exam}\\
\hline
\textbf{\# Fields} &
1 & 2 & 26 & 14 & 8 & 18 & 7 & 43 & 27 & 4 & 37 & 7 \\
\hline
3.1 Pro & \bm{$1.00$}$_{-0.00}^{+0.00}$ & \bm{$0.00$}$_{-0.00}^{+0.00}$ & $0.77_{-0.77}^{+0.23}$ & $0.75_{-0.11}^{+0.13}$ & \bm{$1.00$}$_{-0.29}^{+0.00}$ & \bm{$0.83$}$_{-0.34}^{+0.17}$ & \bm{$1.00$}$_{-0.17}^{+0.00}$ & $0.53_{-0.15}^{+0.17}$ & \bm{$0.96$}$_{-0.12}^{+0.04}$ & \bm{$1.00$}$_{-0.28}^{+0.00}$ & \bm{$0.88$}$_{-0.03}^{+0.13}$ & \bm{$0.67$}$_{-0.07}^{+0.33}$ \\
3.1 Flash Lite & \bm{$1.00$}$_{-0.00}^{+0.00}$ & \bm{$0.00$}$_{-0.00}^{+0.00}$ & $0.71_{-0.71}^{+0.29}$ & \bm{$0.77$}$_{-0.13}^{+0.15}$ & $0.67_{-0.00}^{+0.33}$ & $0.50_{-0.31}^{+0.31}$ & $0.83_{-0.03}^{+0.17}$ & \bm{$0.56$}$_{-0.13}^{+0.13}$ & $0.92_{-0.16}^{+0.08}$ & \bm{$1.00$}$_{-0.33}^{+0.00}$ & $0.86_{-0.15}^{+0.07}$ & $0.60_{-0.20}^{+0.40}$ \\
2.5 Pro & \bm{$1.00$}$_{-0.00}^{+0.00}$ & \bm{$0.00$}$_{-0.00}^{+1.00}$ & $0.75_{-0.42}^{+0.25}$ & $0.71_{-0.10}^{+0.16}$ & \bm{$1.00$}$_{-0.00}^{+0.00}$ & $0.83_{-0.30}^{+0.17}$ & $0.83_{-0.17}^{+0.00}$ & $0.55_{-0.11}^{+0.12}$ & $0.88_{-0.39}^{+0.12}$ & \bm{$1.00$}$_{-0.25}^{+0.00}$ & $0.83_{-0.18}^{+0.10}$ & $0.60_{-0.20}^{+0.40}$ \\
2.5 Flash & \bm{$1.00$}$_{-0.00}^{+0.00}$ & \bm{$0.00$}$_{-0.00}^{+0.76}$ & $0.63_{-0.63}^{+0.38}$ & $0.71_{-0.11}^{+0.14}$ & \bm{$1.00$}$_{-0.33}^{+0.00}$ & $0.67_{-0.20}^{+0.22}$ & $0.67_{-0.05}^{+0.17}$ & $0.50_{-0.15}^{+0.13}$ & $0.91_{-0.11}^{+0.05}$ & \bm{$1.00$}$_{-0.28}^{+0.00}$ & $0.86_{-0.14}^{+0.10}$ & $0.60_{-0.20}^{+0.26}$ \\
GPT 5.4 & \bm{$1.00$}$_{-0.00}^{+0.00}$ & \bm{$0.00$}$_{-0.00}^{+0.76}$ & \bm{$1.00$}$_{-1.00}^{+0.00}$ & $0.71_{-0.14}^{+0.12}$ & \bm{$1.00$}$_{-0.25}^{+0.00}$ & $0.67_{-0.17}^{+0.33}$ & $0.83_{-0.17}^{+0.17}$ & \bm{$0.56$}$_{-0.11}^{+0.13}$ & $0.88_{-0.15}^{+0.08}$ & \bm{$1.00$}$_{-0.25}^{+0.00}$ & $0.79_{-0.15}^{+0.16}$ & $0.60_{-0.20}^{+0.30}$ \\
GPT 5-Mini & \bm{$1.00$}$_{-0.00}^{+0.00}$ & \bm{$0.00$}$_{-0.00}^{+1.00}$ & $0.00_{-0.00}^{+0.71}$ & $0.43_{-0.13}^{+0.20}$ & $0.42_{-0.42}^{+0.33}$ & $0.27_{-0.27}^{+0.23}$ & $0.83_{-0.17}^{+0.00}$ & $0.12_{-0.12}^{+0.19}$ & $0.15_{-0.11}^{+0.42}$ & $0.50_{-0.00}^{+0.25}$ & $0.57_{-0.47}^{+0.28}$ & $0.20_{-0.20}^{+0.32}$ \\
Haiku 4.5 & \bm{$1.00$}$_{-0.00}^{+0.00}$ & \bm{$0.00$}$_{-0.00}^{+0.00}$ & $0.00_{-0.00}^{+0.00}$ & $0.50_{-0.14}^{+0.14}$ & $0.75_{-0.22}^{+0.25}$ & $0.33_{-0.16}^{+0.47}$ & $0.67_{-0.33}^{+0.17}$ & $0.30_{-0.18}^{+0.13}$ & $0.80_{-0.12}^{+0.12}$ & $0.75_{-0.25}^{+0.25}$ & $0.43_{-0.12}^{+0.21}$ & $0.40_{-0.20}^{+0.40}$ \\
Sonnet 4.6 & \bm{$1.00$}$_{-0.00}^{+0.00}$ & \bm{$0.00$}$_{-0.00}^{+0.00}$ & $0.00_{-0.00}^{+0.11}$ & $0.71_{-0.07}^{+0.14}$ & \bm{$1.00$}$_{-0.33}^{+0.00}$ & $0.67_{-0.17}^{+0.33}$ & $0.83_{-0.17}^{+0.00}$ & $0.33_{-0.11}^{+0.13}$ & $0.80_{-0.12}^{+0.14}$ & \bm{$1.00$}$_{-0.25}^{+0.00}$ & $0.79_{-0.12}^{+0.09}$ & $0.63_{-0.19}^{+0.37}$ \\
Opus 4.6 & \bm{$1.00$}$_{-0.00}^{+0.00}$ & \bm{$0.00$}$_{-0.00}^{+0.00}$ & $0.00_{-0.00}^{+0.20}$ & $0.60_{-0.06}^{+0.15}$ & \bm{$1.00$}$_{-0.33}^{+0.00}$ & $0.67_{-0.43}^{+0.16}$ & $0.67_{-0.17}^{+0.13}$ & $0.41_{-0.10}^{+0.09}$ & $0.77_{-0.15}^{+0.13}$ & $0.75_{-0.00}^{+0.25}$ & $0.71_{-0.21}^{+0.04}$ & $0.40_{-0.20}^{+0.27}$ \\
\hline
\end{tabular}}
\renewcommand{\arraystretch}{1.0}
\end{table}
Section specific performance across the real dataset is presented in Table~\ref{tab:median_bounds}. The \textit{Booking Visit} section was excluded from detailed analysis, as it consists of a single, sparsely populated free-text field that proved exceptionally challenging, resulting in uniformly low median performance across all models. Sections dominated by discrete selection fields---\textit{Medical \& General History} and \textit{Future Contraception}---achieve consistently high median accuracies, with leading models frequently exceeding 90\%. This reflects the relative simplicity and standardisation of categorical fields, which are less sensitive to handwriting variability and structural ambiguity.

In contrast, the \textit{Investigations} section emerges as the most challenging, with a maximum median accuracy of only 56\% (3.1 Flash Lite). This difficulty arises from several compounding factors: the frequent inclusion of dates and auxiliary notes outside the intended structure, highly inconsistent and often illegible handwriting due to spatial constraints, and the presence of particularly low-performing fields. Notably, the free text field “other” yielded no correct extractions across any model which highlights the limitations of exact string accuracy with disordered, unconstrained outputs.

Similarly, the \textit{Counselling} section, comprising exclusively date fields, suffers from structural inconsistencies and poor handwriting - making accurate extraction difficult. However, GPT-5.4 stands out as a clear outlier, achieving a perfect median accuracy of 100\%, more than 20 percentage points above the next-best model. This suggests a notable robustness in handling date-specific extraction tasks under challenging conditions.

The \textit{Vaginal Examination} section contains relatively few fields (seven), which contributes to larger performance variation across documents. Additionally, the distribution is positively skewed, indicating there are in majority lower accuracy outcomes with occasional high performing instances. 

The final section, \textit{Obstetric \& Neonatal History} consists of a mix of formatted fields, discrete selection fields, and normalisation requirements. 
Gemini 3.1 Pro achieves the strongest performance in this section, with a median accuracy of 88\%, indicating superior capability in handling semantically complex fields.

\begin{table}[H]
\centering
\caption{Top 4 best, worst, and most hallucinated fields across the real dataset with the optimal prompt and template. Averaged across all models excluding 5-Mini and Haiku; only fields with $n \ge 5$ filled fields are included within the dataset are included. Most hallucinated fields is defined by the hallucination rate. Rates are shown in brackets.}
\label{tab:top4_fields}
\renewcommand{\arraystretch}{1.6}
\resizebox{0.85\textwidth}{!}{
\begin{tabular}{|l|l|l|}
\hline
\textbf{Best Performing Fields (w/o nulls)} & 
\textbf{Worst Performing Fields (w/o nulls)} & 
\textbf{Most Hallucinated Fields} \\
\hline
Discussed (0.98)           & Other (0.00)                                & EDD method (1.0)              \\
Outcome (0.95)             & Screening for gestational diabetes 2 (0.00) & Family History Diabetes (0.98) \\
Discussed in record (0.95) & Tetox notes (0.03)                          & Family history TB (0.98)       \\
Permission obtained (0.92) & Screening for gestational diabetes (0.06)   & HIV status at booking (0.96)   \\
\hline
\end{tabular}}
\renewcommand{\arraystretch}{1.0}
\end{table}

The worst-performing and most frequently hallucinated fields (Table~\ref{tab:top4_fields}) provide important insight into the specific data structures and field types that remain challenging for current models. A particularly notable example is the EDD methods fields, which exhibit a high hallucination rate. This appears to stem from systematic misinterpretation of the checkbox structure, where models incorrectly infers a selected option despite the absence of a clear mark, effectively treating the visual layout itself as a signal.

Another unexpectedly poor performing category is the screening for gestational diabetes fields. In principle, these should be relatively straightforward, requiring only the extraction of a numerical value and corresponding unit (e.g., mmol, mmol/L, HGT). However, variability in unit representation across documents and particularly poor handwriting within this field resulted in ambiguity for the models and overall poor performance. This suggests that model performance in such fields is sensitive to a lack of standardisation and would likely benefit from downstream normalisation during post-processing. Furthermore, models frequently default to using full stops rather than commas as decimal separators, irrespective of the notation present in the source document, indicating a bias toward formatting conventions over transcription.

The HIV status at booking field also demonstrates relatively poor performance, despite being a simple categorical variable. The underlying cause is less immediately clear, suggesting that errors may arise from a combination of OCR limitations, ambiguous handwriting, or inconsistencies in annotation. 

The free-text “other” field within the \textit{Investigations} section predictably yields extremely low exact match accuracy, with no correct predictions across all models. This field represents one of the most unstructured regions of the form: entries are often written outside designated boundaries, are lengthy, and contain a mixture of textual and numerical information in highly variable formats. Although model outputs are frequently semantically close to the ground truth, minor discrepancies---omitted words, missing numerical values, or slight formatting differences (e.g., exclusion of degree symbols)---result in incorrect classifications under strict exact-match evaluation. This highlights a key limitation of both the models' transcription capabilities for complex free text strings and exact string matching as an evaluation metric for free text extraction.

In contrast, the best performing fields are in majority discrete selection fields. The "outcome" fields, in particular, achieves consistently high accuracy despite containing visually similar options. This indicates that, for structured selection tasks, OCR limitations are not the primary bottleneck; rather, models are generally capable of reliably distinguishing between clearly defined discrete options.
\subsection{Ablation Study: Prompt Structure}
\label{Ablation}
\begin{table}[H]
\centering
\scriptsize
\caption{Base (ablated) prompt  performance metrics on the real-world document dataset. Metrics include median values with 16th and 84th percentiles; word error rate (WER) and character error rate (CER) measure the ratio of words and characters respectively---CER and WER were only calculated for free text fields. Hallucination rate indicates proportion of incorrect or fabricated outputs. Gemini 3.1 remains the best performing model, suprisingly the WER decreases for some models with this simplified prompt however the CER becomes substantially higher and overall performance suffers a noticeable decrease across most models. Budget models such as Haiku 4.5 and GPT 5-Mini experience a performance increase due to the models not being able to interpret the previous discretised template and returning the list of options rather than choosing the correct one.}
\label{tab:metrics_median_bounds}
\resizebox{0.8\textwidth}{!}{
\renewcommand{\arraystretch}{1.6}
\begin{tabular}{|l| r| r r r r| r r| r r|}
\hline
\shortstack{Model} & 
\shortstack{Docs} &
\shortstack{Accuracy} &
\shortstack{Accuracy\\(w/o nulls)} &
\shortstack{Formatted\\ Accuracy} &
\shortstack{Formatted\\Accuracy\\(w/o nulls)} &
\shortstack{WER} & 
\shortstack{CER} & 
\shortstack{Null Field\\Accuracy} &
\shortstack{Hallucination\\Rate}\\
\hline
3.1 Pro       & 49 & \bm{$0.84$}$_{-0.08}^{+0.05}$ & \bm{$0.78$}$_{-0.23}^{+0.06}$ & $0.88_{-0.13}^{+0.08}$ & $0.76_{-0.35}^{+0.20}$ & \textbf{0.34} & 0.92 & 0.08 & 0.08 \\
3.1 Flash Lite     & 49 & $0.80_{-0.04}^{+0.06}$ & $0.68_{-0.01}^{+0.05}$ & $0.85_{-0.08}^{+0.08}$ & $0.71_{-0.14}^{+0.21}$ & 0.43 & 0.92 & 0.08 & 0.08 \\
2.5 Pro      & 48 & $0.78_{-0.08}^{+0.06}$ & $0.73_{-0.13}^{+0.03}$ & $0.86_{-0.07}^{+0.08}$ & \bm{$0.79$}$_{-0.14}^{+0.09}$ & 0.62 & 0.42 & 0.84 & 0.16 \\
2.5 Flash    & 49 & $0.78_{-0.06}^{+0.06}$ & $0.66_{-0.06}^{+0.09}$ & $0.82_{-0.07}^{+0.07}$ & $0.65_{-0.16}^{+0.15}$ & 0.62 & 0.46 & \textbf{0.88} & 0.12 \\
GPT 5.4      & 48 & \bm{$0.84$}$_{-0.04}^{+0.04}$ & $0.73_{-0.07}^{+0.06}$ & \bm{$0.90$}$_{-0.08}^{+0.04}$ & \bm{$0.79$}$_{-0.15}^{+0.15}$ & 0.44 & 0.94 & 0.06 & \textbf{0.06} \\
GPT 5-Mini    & 49 & $0.71_{-0.07}^{+0.13}$ & $0.49_{-0.01}^{+0.13}$ & $0.75_{-0.09}^{+0.08}$ & $0.39_{-0.17}^{+0.19}$ & 0.61 & 0.93 & 0.07 & 0.07 \\
Haiku 4.5        & 46 & $0.59_{-0.07}^{+0.06}$ & $0.41_{-0.01}^{+0.09}$ & $0.63_{-0.11}^{+0.12}$ & $0.25_{-0.07}^{+0.13}$ & 0.75 & \textbf{0.25} & 0.75 & 0.25 \\
Sonnet 4.6& 46 & $0.74_{-0.05}^{+0.06}$ & $0.58_{-0.01}^{+0.06}$ & $0.73_{-0.07}^{+0.09}$ & $0.38_{-0.11}^{+0.15}$ & 0.39 & 0.89 & 0.11 & 0.11 \\
Opus 4.6         & 46 & $0.71_{-0.04}^{+0.08}$ & $0.53_{-0.02}^{+0.05}$ & $0.69_{-0.08}^{+0.10}$ & $0.35_{-0.15}^{+0.17}$ & 0.58 & 0.90 & 0.10 & 0.10 \\
\hline
\end{tabular}}
\renewcommand{\arraystretch}{1.0}
\end{table}
\begin{table}[H]
\centering
\scriptsize
\caption{Base (ablated) prompt performance on the real-world document dataset showing precision, recall and F1 on a macro and weighted level. The last column shows the drop in weighted and macro F1 score respectively compared to the optimised prompt results. This is indicative of the model performing worse overall with the ablated prompt in discrete selection fields particularly surrounding uncommon selections.}
\label{tab:metrics_base}

\fontsize{5pt}{7pt}\selectfont

\resizebox{0.8\textwidth}{!}{%
\renewcommand{\arraystretch}{1}
\begin{tabular}{|l| r r r r r r| r r|}
\hline
\shortstack{Model} &
\shortstack{Macro\\Precision} &
\shortstack{Macro\\Recall} &
\shortstack{Macro\\F1} &
\shortstack{Weighted\\Precision} &
\shortstack{Weighted\\Recall} &
\shortstack{Weighted\\ F1} &
\shortstack{$\Delta$ Macro\\ F1} &
\shortstack{$\Delta$ Weighted\\ F1} \\
\hline
3.1 Pro      & \textbf{0.41} & \textbf{0.39} & \textbf{0.39} & 0.87 & \textbf{0.86} & 0.86 &-0.27 & -0.05 \\
3.1 Flash Lite   & 0.32 & 0.28 & 0.28 & 0.83 & 0.82 & 0.82 &-0.28 & -0.05 \\
2.5 Pro     & 0.24 & 0.23 & 0.22 & 0.86 & 0.78 & 0.80 &-0.24 & -0.05 \\
2.5 Flash   & 0.32 & 0.31 & 0.30 & 0.84 & 0.81 & 0.82 & -0.21& -0.05 \\
GPT 5.4     & 0.32 & 0.31 & 0.30 & \textbf{0.88} & \textbf{0.86} & \textbf{0.87}&-0.26 & -0.02 \\
GPT 5-Mini   & 0.29 & 0.22 & 0.24 & 0.72 & 0.66 & 0.66 & 0.09 & 0.22 \\
Haiku 4.5       & 0.22 & 0.18 & 0.18 & 0.72 & 0.60 & 0.63 &-0.16 & -0.12 \\
Sonnet 4.6      & 0.29 & 0.26 & 0.25 & 0.83 & 0.78 & 0.79 &-0.20 & -0.05 \\
Opus 4.6        & 0.20 & 0.18 & 0.17 & 0.81 & 0.78 & 0.79 &-0.29 & -0.04 \\
\hline
\end{tabular}%
}
\end{table}
Prompt ablation in this study primarily focused on modifications to the extraction template rather than the instruction prompt itself. In the optimal configuration, the template was discretised to include explicit enumerations of valid responses for all discrete selection fields, whereas the prompt consisted predominantly of normalisation rules. Overall extraction accuracy only increased modestly between the optimised and ablated prompt, approximately 1-3\% across the best performing models (Table ~\ref{tab:metrics_median_bounds}). This is expected as the discretisation was only applied to a small subset of fields within the form. Considering the performance change within discrete fields only, we can see a percentage drop across macro precision, recall and F1 score of $\sim{50}\%$ across all leading models (Table~\ref{tab:metrics_base}). This is indicative that the model is creating substantially more classifications for the various discrete fields. The weighted metrics experience substantially smaller performance, with weighted F1 score dropping 2-5\% drops indicating that the overall performance remains strong. In rare cases or particularly difficult forms, the model performance deteriorates without discretisation. Gemini 3.1 Pro also experienced a substantially higher truncation rate with 8 documents failing to produce a full output with the base prompt compared to the 2 documents in the optimised prompt. This is evidence of the stability and performance value additional discretisation and information can provide to LLMs models when applicable. Interestingly, smaller models such as GPT-5 Mini and Haiku 4.5 performed substantially better under the base prompt configuration. Without the discretised template, these models relied primarily on their visual recognition capabilities rather than attempting to interpret structured reasoning constraints. This suggests that older models and models with lower inference and reasoning capabilities may struggle with schema-driven extraction tasks that require selecting among predefined categories.


\section{Conclusions} 
The digitisation of handwritten medical information represents a bottleneck in the healthcare data infrastructure, especially in resource-constrained settings. We benchmarked a comprehensive set of 17 frontier and open-source multimodal models against synthetic and real handwritten Maternity Case Records provided by the UBOMI BUHLE Pregnancy Exposure Registry, evaluating performance across discrete selection, format constrained and free-text fields. We found that the best LLMs did surprisingly well, achieving a median of approximately 85\% for exact string match accuracy across all forms with Gemini 3.1 Pro followed by GPT 5.4 being the strongest performing models. Free text fields were---as expected---the weakest performing field structure across the data set with the best model being Gemini 3.1 Pro, which in this area achieved a 30\% CER and 50\% WER. Gemini 3.1 Pro was additionally the best performing model across discrete selection fields achieving the best performance across all classification metrics with a weighted F1 score of 0.91. GPT 5.4 excelled in noisy date extraction achieving a 100\% median accuracy for the counselling section. It also performed marginally better than other models in minimising hallucinations achieving a 6\% hallucination rate. For formatted fields, Claude Sonnet leads with 77\% median accuracy (excluding nulls) followed by Haiku 4.5 (73\%), Gemini 3.1 Flash Lite (72\%) and GPT-5 Mini (72\%); indicating that model size and reasoning capabilities do not necessarily benefit fields with implicit structure (dates and numerical values). We found that the small open-source models ($< 5$B parameters) we considered did not perform well. However, we emphasise that this should not be taken as indicative of all open-source vision models, but rather only those limited by standard consumer T4 GPU memory.

Though our dataset was relatively small (with 49 forms), the heterogeneity and complexity of the form provides an excellent and challenging benchmark for a wide range of digitisation challenges, including bad handwriting, dates, abbreviations and writing outside lines/at angles that are difficult even for humans untrained in maternal healthcare to understand. The results provide strong evidence that we have reached the point where multimodal large language models constitute a promising foundation for reliable and scalable document digitisation, potentially unlocking vast archives of previously inaccessible medical and administrative data, creating new opportunities for research, improved healthcare delivery, and broader societal benefit. Future work could involve increasing the dataset, helping to mitigate sparsity effects and expanding to various different forms.

\section*{Acknowledgements}

This study was supported by the UBOMI BUHLE Pregnancy Exposure registry project through funding from the Gates Foundation (INV-004508). The findings and conclusions in this publication are those of the author(s) and do not necessarily represent the official position of the funding agency. We thank Linda Camara for initial work related to this project and the UBOMI BUHLE team based at ESRU, Rahima Moosa Mother and Child Hospital for capturing, curating and redacting the data and creating the realistic synthetic data.

\bibliographystyle{unsrt}  
\bibliography{references}  

@misc{crosilla2025benchmarkinglargelanguagemodels,
      title={Benchmarking Large Language Models for Handwritten Text Recognition}, 
      author={Giorgia Crosilla and Lukas Klic and Giovanni Colavizza},
      year={2025},
      eprint={2503.15195},
      archivePrefix={arXiv},
      primaryClass={cs.CV},
      note={arXiv:2503.15195},
      url={https://arxiv.org/abs/2503.15195}, 
}

@misc{bai2025qwen25vltechnicalreport,
      title={Qwen2.5-VL Technical Report}, 
      author={Shuai Bai and Keqin Chen and Xuejing Liu and Jialin Wang and
              Wenbin Ge and Sibo Song and Kai Dang and Peng Wang and Shijie Wang and
              Jun Tang and Humen Zhong and Yuanzhi Zhu and Mingkun Yang and
              Zhaohai Li and Jianqiang Wan and Pengfei Wang and Wei Ding and
              Zheren Fu and Yiheng Xu and Jiabo Ye and Xi Zhang and Tianbao Xie and
              Zesen Cheng and Hang Zhang and Zhibo Yang and Haiyang Xu and Junyang Lin},
      year={2025},
      eprint={2502.13923},
      archivePrefix={arXiv},
      primaryClass={cs.CV},
      note={arXiv:2502.13923},
      url={https://arxiv.org/abs/2502.13923}, 
}

@misc{li2025dotsocrmultilingualdocumentlayout,
      title={dots.ocr: Multilingual Document Layout Parsing in a Single Vision-Language Model}, 
      author={Yumeng Li and Guang Yang and Hao Liu and Bowen Wang and Colin Zhang},
      year={2025},
      eprint={2512.02498},
      archivePrefix={arXiv},
      primaryClass={cs.CV},
      note={arXiv:2512.02498},
      url={https://arxiv.org/abs/2512.02498}, 
}

@misc{shi2023exploringocrcapabilitiesgpt4vision,
      title={Exploring OCR Capabilities of GPT-4V(ision): A Quantitative and In-depth Evaluation}, 
      author={Yongxin Shi and Dezhi Peng and Wenhui Liao and Zening Lin and Xinhong Chen and
              Chongyu Liu and Yuyi Zhang and Lianwen Jin},
      year={2023},
      eprint={2310.16809},
      archivePrefix={arXiv},
      primaryClass={cs.CV},
      note={arXiv:2310.16809},
      url={https://arxiv.org/abs/2310.16809}, 
}

@misc{li2022trocrtransformerbasedopticalcharacter,
      title={TrOCR: Transformer-based Optical Character Recognition with Pre-trained Models}, 
      author={Minghao Li and Tengchao Lv and Jingye Chen and Lei Cui and Yijuan Lu and
              Dinei Florencio and Cha Zhang and Zhoujun Li and Furu Wei},
      year={2022},
      eprint={2109.10282},
      archivePrefix={arXiv},
      primaryClass={cs.CL},
      note={arXiv:2109.10282},
      url={https://arxiv.org/abs/2109.10282}, 
}

@misc{lee2023pix2structscreenshotparsingpretraining,
      title={Pix2Struct: Screenshot Parsing as Pretraining for Visual Language Understanding}, 
      author={Kenton Lee and Mandar Joshi and Iulia Turc and Hexiang Hu and Fangyu Liu and
              Julian Eisenschlos and Urvashi Khandelwal and Peter Shaw and
              Ming-Wei Chang and Kristina Toutanova},
      year={2023},
      eprint={2210.03347},
      archivePrefix={arXiv},
      primaryClass={cs.CL},
      note={arXiv:2210.03347},
      url={https://arxiv.org/abs/2210.03347}, 
}

@misc{layoutLM,
      title={LayoutLM: Pre-training of Text and Layout for Document Image Understanding},
      author={Yiheng Xu and Minghao Li and Lei Cui and Shaohan Huang and Furu Wei and Ming Zhou},
      year={2019},
      eprint={1912.13318},
      archivePrefix={arXiv},
      primaryClass={cs.CL},
      note={arXiv:1912.13318},
      url={https://arxiv.org/abs/1912.13318}
}

@inproceedings{Lopresti_ocr_errors,
  author    = {Lopresti, Daniel},
  title     = {Optical character recognition errors and their effects on natural language processing},
  year      = {2008},
  publisher = {Association for Computing Machinery},
  address   = {New York, NY, USA},
  url       = {https://doi.org/10.1145/1390749.1390753},
  doi       = {10.1145/1390749.1390753},
  booktitle = {Proceedings of the Second Workshop on Analytics for Noisy Unstructured Text Data},
  pages     = {9--16}
}

@inproceedings{kumar_interpreting_doctor_notes,
  author    = {Kumar, Nithin and Chikkmath, Amrutha and B R, Gouthami Yadav and
               Naidu, Hema R and Acharya, Diya},
  booktitle = {2023 International Conference on Advances in Electronics, Communication,
               Computing and Intelligent Information Systems (ICAECIS)}, 
  title     = {Interpreting Doctor notes using Handwriting Recognition and
               Deep Learning Techniques: A Survey}, 
  year      = {2023},
  pages     = {703--708},
  doi       = {10.1109/ICAECIS58353.2023.10170398}
}

@article{2021arXiv211115664K,
  author        = {{Kim}, Geewook and {Hong}, Teakgyu and {Yim}, Moonbin and
                   {Nam}, Jeongyeon and {Park}, Jinyoung and {Yim}, Jinyeong and
                   {Hwang}, Wonseok and {Yun}, Sangdoo and {Han}, Dongyoon and
                   {Park}, Seunghyun},
  title         = {{OCR-free Document Understanding Transformer}},
  journal       = {arXiv e-prints},
  year          = 2021,
  month         = nov,
  pages         = {arXiv:2111.15664},
  doi           = {10.48550/arXiv.2111.15664},
  archivePrefix = {arXiv},
  eprint        = {2111.15664},
  primaryClass  = {cs.LG},
  note          = {arXiv:2111.15664}
}

@article{wang2024qwen2vl,
  title   = {Qwen2-{VL}: Enhancing Vision-Language Model's Perception of the
             World at Any Resolution},
  author  = {Wang, Peng and Bai, Shuai and Tan, Sinan and Wang, Shijie and
             Fan, Zhihao and Bai, Jinze and Chen, Keqin and Liu, Xuejing and
             Wang, Jialin and Ge, Wenbin and Fan, Yang and Dang, Kai and
             Du, Mengfei and Ren, Xuancheng and Men, Rui and Liu, Dayiheng and
             Zhou, Chang and Zhou, Jingren and Lin, Junyang},
  journal = {arXiv preprint arXiv:2409.12191},
  year    = {2024},
  note    = {arXiv:2409.12191},
  url     = {https://arxiv.org/abs/2409.12191}
}

@misc{achiam2023gpt4,
      title={GPT-4 Technical Report},
      author={OpenAI and Josh Achiam and Steven Adler and Sandhini Agarwal and
              Lama Ahmad and Ilge Akkaya and Florencia Leoni Aleman and
              Diogo Almeida and Janko Altenschmidt and Sam Altman and
              Shyamal Anadkat and Red Avila and Igor Babuschkin and
              Suchir Balaji and Valerie Balcom and Paul Baltescu and
              Haiming Bao and Mohammad Bavarian and Jeff Belgum and
              Irwan Bello and Jake Berdine and Greg Brockman and others},
      year={2023},
      eprint={2303.08774},
      archivePrefix={arXiv},
      primaryClass={cs.CL},
      note={arXiv:2303.08774},
      url={https://arxiv.org/abs/2303.08774},
}

@misc{geminiteam2025gemini25,
  title         = {Gemini 2.5: Pushing the Frontier with Advanced Reasoning,
                   Multimodality, Long Context, and Next Generation Agentic Capabilities},
  author        = {{Gemini Team} and Comanici, Gheorghe and others},
  year          = {2025},
  eprint        = {2507.06261},
  archivePrefix = {arXiv},
  primaryClass  = {cs.AI},
  note          = {arXiv:2507.06261},
  url           = {https://arxiv.org/abs/2507.06261}
}

@misc{anthropic2025haiku,
  title       = {Claude Haiku 4.5},
  author      = {{Anthropic}},
  institution = {Anthropic},
  year        = {2025},
  month       = oct,
  note        = {\url{https://www.anthropic.com/claude/haiku}}
}

@misc{anthropic2026opus,
  title       = {Claude Opus 4.6},
  author      = {{Anthropic}},
  institution = {Anthropic},
  year        = {2026},
  month       = feb,
  note        = {\url{https://www.anthropic.com/claude/opus}}
}

@misc{anthropic2025sonnet,
  title       = {Claude Sonnet 4.6},
  author      = {{Anthropic}},
  institution = {Anthropic},
  year        = {2025},
  month       = sep,
  note        = {\url{https://www.anthropic.com/claude/sonnet}}
}

@misc{google2025gemini31pro,
  title       = {Gemini 3.1 Pro},
  author      = {{Google DeepMind}},
  institution = {Google DeepMind},
  year        = {2025},
  note        = {Preview model. \url{https://deepmind.google/models/gemini/pro/}}
}

@techreport{google2025gemini31flashlite,
  title       = {Gemini 3.1 {Flash-Lite}},
  author      = {{Google DeepMind}},
  institution = {Google DeepMind},
  year        = {2025},
  note        = {\url{https://storage.googleapis.com/deepmind-media/Model-Cards/Gemini-3-1-Flash-Lite-Model-Card.pdf}}
}

@misc{openai2025gpt41,
  title       = {{GPT}-4.1},
  author      = {{OpenAI}},
  institution = {OpenAI},
  year        = {2025},
  note        = {\url{https://openai.com/index/gpt-4-1/}}
}

@misc{openai2025gpt5,
      title={{OpenAI} {GPT}-5 System Card},
      author={Aaditya Singh and Adam Fry and Adam Perelman and Adam Tart and others},
      year={2025},
      eprint={2601.03267},
      archivePrefix={arXiv},
      primaryClass={cs.CL},
      note={arXiv:2601.03267},
      url={https://arxiv.org/abs/2601.03267}
}

@misc{xai2025grok41,
  title       = {Grok 4.1 Model Card},
  author      = {{xAI}},
  institution = {xAI},
  year        = {2025},
  note        = {\url{https://data.x.ai/2025-11-17-grok-4-1-model-card.pdf}}
}

\appendix

\section{Extraction Prompt}
\label{app:prompt}

Here we show the prompt used to extract text from both the real and synthetic data:  

\begin{tcolorbox}[
  breakable,
  enhanced,
  fonttitle=\bfseries,
  colback=gray!5,
  colframe=gray!60,
  fontupper=\ttfamily\footnotesize
]
You are an expert medical form data extractor specializing in the South African
Maternity Case Record (MCR1). You are required to digitise the images into the
template format provided in the separate template .json file.

\textbf{Hard rules:}
\begin{itemize}
  \item Use double quotes for ALL keys and ALL string values.
  \item Do NOT include trailing commas anywhere.
  \item Do NOT include comments, explanations, or markdown.
\end{itemize}

\textbf{GENERAL RULES:}
\begin{itemize}
  \item Extract ALL visible handwriting.
  \item Do NOT correct spelling (unless normalizing specific locations or
        abbreviations defined below).
  \item Do NOT infer missing information.
  \item \textbf{Strip Leading Zeros:} For integer fields (Gravida, Para, Age,
        etc.), convert ``04'' to ``4''.
        EXCEPTION: Do NOT strip zeros from Date parts
        (e.g., maintain ``01'' in ``01/05/2024'').
  \item \textbf{Underscore Neutrality:} Treat values with underscores as
        equivalent to their plain versions (e.g., ``Yes\_'', ``\_Yes'', or
        ``Yes'' should all be output as ``Yes''). This applies to ``Yes'' and
        ``No'' across all sections.
  \item ``\textdegree{}'' means ``no'' --- e.g., ``\textdegree{} APH''
        $\rightarrow$ ``no APH''.
\end{itemize}

\textbf{SPECIFIC FORMATTING RULES:}
\begin{itemize}
  \item Dates: Extract wherever written (inside or outside boxes).
\end{itemize}

\textbf{SPECIFIC NORMALIZATION RULES:}

\textit{1. Abbreviation \& Status Normalization:}
\begin{itemize}
  \item \textbf{Sex:} ``Male'' $\rightarrow$ ``M'', ``Female'' $\rightarrow$ ``F''.
  \item \textbf{Outcome:} ``Alive'' $\rightarrow$ ``A'',
        ``Intra-uterine death'' $\rightarrow$ ``IUD'',
        ``Neonatal death'' $\rightarrow$ ``NND'',
        ``Infant death'' $\rightarrow$ ``ID''.
  \item ``neg'' $\rightarrow$ ``Negative'', ``pos'' $\rightarrow$ ``Positive''.
\end{itemize}

\textit{3. Date Formatting:}
\begin{itemize}
  \item \textbf{Output Format:} ALWAYS use \textbf{DD/MM/YYYY}.
  \item \textbf{Year Expansion:} Expand 2-digit years to 4 digits
        (e.g., ``24'' $\rightarrow$ ``2024'').
  \item \textbf{Zero Padding:} Ensure days and months are zero-padded
        (e.g., ``01/05/2024'').
\end{itemize}

\textit{4. Gestation/Duration Formatting (Obstetric \& Neonatal History only):}
\begin{itemize}
  \item ``9 weeks'' or ``9/52'' $\rightarrow$ ``9/40''.
  \item Normalise ``9/12'' $\rightarrow$ ``9 months''.
  \item Do NOT normalise ``Term'' or ``term'' --- write verbatim if present.
\end{itemize}

\textit{5. Blood Pressure:} Extract \texttt{bp} (systolic) and
\texttt{bp\_dia} (diastolic) separately.
Example: ``120/70'' $\rightarrow$ \texttt{bp}: ``120'',
\texttt{bp\_dia}: ``70''.

\textit{6. Medical \& General History} --- For use of herbal, use of otc,
tobacco\_use, alcohol\_use, substance\_use:
``1'' = User, ``2'' = Former User, ``3'' = Non-User.

\textit{7.} For the following specific fields use ``1'' for
Checked/Ticked/Crossed/Filled/Yes and ``0'' for Empty/No:
\texttt{hypertension, diabetes, cardiac, asthma, tuberculosis, epilepsy,
mental\_health\_disorder, hiv, other\_condition, none, edd\_method\_lnmp,
edd\_method\_sf, edd\_method\_sonar, family\_history\_tb,
family\_history\_twins, family\_history\_diabetes}.

\textit{8. All Other Checkboxes (Yes/null):}
\begin{itemize}
  \item If Ticked/Checked/Crossed/Filled/Yes $\rightarrow$ ``Yes''.
  \item If Empty/Blank $\rightarrow$ null.
\end{itemize}

\textit{9. Medical \& General History:}
\begin{itemize}
  \item ``none'' is a summary field: ``1'' if ALL personal condition fields
        were negative, otherwise ``0''. Does NOT include family history fields.
  \item ``allergies'' is often written ``Nil known'', ``nil known'', ``none''.
  \item \texttt{tb\_symptom\_screen} options: ``Positive'' and ``Negative''.
\end{itemize}

\textit{10. Investigations:}
\begin{itemize}
  \item \texttt{hiv\_status\_at\_booking}: ``Un-known'', ``Positive'',
        ``Negative''.
  \item \texttt{rhesus} and \texttt{hiv\_test[1-3]\_result}: ``Positive'',
        ``Negative'' only.
  \item HIV Retest: only option is ``NotDecli'', otherwise leave blank.
        If the retest field is filled then ``NotDecli'' MUST be written.
\end{itemize}

\textit{11.} Syphilis test results: write only the numerical value:
``1'' = Positive, ``2'' = Negative.

\textit{12.} ``? xxx'' --- copy verbatim with the question mark.

\textit{13.} For sections with space constraints (particularly
\textit{Investigations}), information may be written outside the block space.
Attempt to capture this data where applicable.

\textit{14. Gestational Age:}
\begin{itemize}
  \item Fields \texttt{bpd, hc, ac, fl, average\_gestation}: format
        ``xxwksxdys''.
  \item Field \texttt{crl}: weight must be written in grams.
\end{itemize}

\textit{15.} \texttt{overwrite} should only be filled with ``Primigravida''
if the first pregnancy has been written across all rows of obstetric and
neonatal history; otherwise null.

\textit{16. Vaginal Investigations:} Common answers include combinations of
``no sores'', ``no warts'', ``no foul smelling discharge'',
``no abnormal discharge'', ``NAD'', ``no bleeding'' --- copy verbatim.

\textit{17. Examinations} --- \texttt{urinalysis\_protein} and
\texttt{urinalysis\_glucose}:
\begin{itemize}
  \item ``\textdegree{}prot, \textdegree{}gluc'' or
        ``pHX \textdegree{}prot, \textdegree{}gluc'' $\rightarrow$
        ``None'' for both fields.
  \item ``No protein, no glucose''; ``No glucose, no protein'';
        ``NAD'' \& ``NAD/POS'' $\rightarrow$ ``None'' for both fields.
  \item Discard information unrelated to protein and glucose.
\end{itemize}

\textbf{EXTRACTION RULES:}
\begin{itemize}
  \item Use EXACT field names from the JSON provided within the template.
  \item Extract all visible handwritten and printed data.
  \item Return ONLY valid JSON.
\end{itemize}
\end{tcolorbox}

\section{JSON Output Template}
\label{app:template}

Below we show the output template that distinguished the optimised prompt from the default prompt. In the latter case the json file was the same but all values were empty. 

\begin{tcblisting}{
  breakable,
  enhanced,
  fonttitle=\bfseries,
  colback=gray!5,
  colframe=gray!60,
  listing only,
  listing options={
    basicstyle=\ttfamily\footnotesize,
    breaklines=true,
    columns=fullflexible
  }
}
{
  "form_name": "Maternity Case Record (MCR1)",
  "sections": [
    {
      "section_name": "Header Identification",
      "fields": {
        "admire_id": null,
        "healthcare_worker_name": null,
        "clinic": null,
        "clinic_date": null,
        "folder_number": null,
        "patient_name": null,
        "age": null,
        "gravida": null,
        "para": null,
        "miscarriages": null
      }
    },
    {
      "section_name": "Obstetric and Neonatal History",
      "records": [
        {
          "year_[1-5]": null,
          "gestation_[1-5]": null,
          "delivery_[1-5]": ["NVD","NV","C/S","C/Sec","Miscarriage","Misc",
                         "Mis","Termination of pregnancy","TOP",
                         "Ectopic Pregnancy"],
          "weight_[1-5]": null,
          "sex_[1-5]": ["M","F"],
          "outcome_[1-5]": ["A","IUD","NND","ID"],
          "complications_[1-5]": null
        }
      ],
      "description_of_complications": null,
      "overwrite": null
    },
    {
      "section_name": "Medical and General History",
      "fields": {
        "hypertension": ["0","1"],
        "diabetes": ["0","1"],
        "cardiac": ["0","1"],
        "asthma": ["0","1"],
        "tuberculosis": ["0","1"],
        "epilepsy": ["0","1"],
        "mental_health_disorder": ["0","1"],
        "hiv": ["0","1"],
        "other_condition": ["0","1"],
        "other_condition_detail": null,
        "none": ["0","1"],
        "family_history_twins": ["0","1"],
        "family_history_diabetes": ["0","1"],
        "family_history_tb": ["0","1"],
        "family_history_congenital": ["0","1"],
        "family_history_details": null,
        "medication": null,
        "operations": null,
        "allergies": null,
        "tb_symptom_screen": ["Positive","Negative"],
        "use_of_herbal": ["1","2","3"],
        "use_of_otc": ["1","2","3"],
        "tobacco_use": ["1","2","3"],
        "alcohol_use": ["1","2","3"],
        "substance_use": ["1","2","3"],
        "type_of_substance_used": null,
        "psychosocial_risk_factors": null
      }
    },
    {
      "section_name": "Examination",
      "fields": {
        "bp": null,
        "bp_dia": null,
        "urinalysis_glucose": ["None","trace","++"],
        "urinalysis_protein": ["None","trace","++"],
        "height": null,
        "weight": null,
        "muac": null,
        "bmi": null,
        "thyroid": ["not enlarged","swollen","not swollen","NAD"],
        "breasts": ["soft","soft,no lumps"],
        "heart": null,
        "lungs": null,
        "abdomen": null,
        "sf_measurement_at_booking": null
      }
    },
    {
      "section_name": "Vaginal Examination",
      "fields": {
        "permission_obtained": ["Yes","No"],
        "vulva_and_vagina": null,
        "cervix": null,
        "uterus": null,
        "pap_smear_done": ["Yes","No"],
        "pap_smear_date": null,
        "pap_smear_result": null
      }
    },
    {
      "section_name": "Investigations",
      "fields": {
        "syphilis_test_date": null,
        "syphilis_test_result": ["1","2"],
        "repeat_syphilis_test_date": null,
        "repeat_syphilis_test_result": ["1","2"],
        "treatment_[1-3]": null,
        "rhesus": ["Positive","Negative"],
        "antibodies": null,
        "hb": null,
        "urine_mcs_date": null,
        "urine_mcs_result": null,
        "tet_tox_[1-3]": null,
        "tetox_notes": null,
        "screening_for_gestational_diabet": null,
        "screening_for_gestational_diabe2": null,
        "screening_gdm_28w": null,
        "hiv_status_at_booking": ["Un-known","Positive","Negative"],
        "hiv_booking_date": null,
        "hiv_booking_result": ["Positive","Negative"],
        "hiv_booking_on_art": ["Yes","No"],
        "hiv_retest_[1-3]_date": null,
        "hiv_retest_[1-3]_result": ["Positive","Negative"],
        "hiv_retest_[1-3]_declined": null,
        "cd4": null,
        "art_initiated_on": null,
        "art_y": null,
        "art_m": null,
        "viral_load_[1-3]_date": null,
        "viral_load_[1-3]_result": null,
        "other": null
      }
    },
    {
      "section_name": "Gestational Age",
      "fields": {
        "lnmp": null,
        "certain": ["1","2"],
        "sonar_date": null,
        "bpd": null,
        "hc": null,
        "ac": null,
        "fl": null,
        "crl": null,
        "placenta": ["Anterior_","Posterior"],
        "afi": null,
        "average_gestation": null,
        "singleton": null,
        "multiple_pregnancy": null,
        "intrauterine_pregnancy": null,
        "estimated_date_of_delivery": null,
        "edd_method_sonar": ["0","1"],
        "edd_method_sf": ["0","1"],
        "edd_method_lnmp": ["0","1"]
      }
    },
    {
      "section_name": "Mental Health",
      "fields": {
        "screening_performed": ["Yes","No"],
        "mental_health_score": null,
        "discussed_in_record": null,
        "referred_to": null
      }
    },
    {
      "section_name": "Birth Companion",
      "fields": {
        "discussed": null
      }
    },
    {
      "section_name": "Counselling",
      "fields": {
        "counselling_[1-13]_date_[1-2]": null
      }
    },
    {
      "section_name": "Future Contraception",
      "fields": {
        "implant": null,
        "inject": null,
        "iud": null,
        "tubal_ligation": null,
        "oral": null,
        "management_plans_discussed": null,
        "educational_material_given": null,
        "tubal_ligation_counselling": null
      }
    },
    {
      "section_name": "Booking Visit and Assessment",
      "fields": {
        "done_by": null,
        "date": null
      }
    }
  ]
}
\end{tcblisting}

\end{document}